\documentclass[times, review, 10pt]{elsarticle}
\usepackage{amssymb}
\usepackage{graphicx}
\usepackage{amsmath}
\usepackage{longtable}
\usepackage{a4wide}
\usepackage{algorithm} 
\usepackage{algpseudocode} 
\usepackage{algorithm}
\usepackage{algpseudocode}
\usepackage{mathrsfs}
\usepackage{subcaption}
\usepackage{mathtools}
\usepackage{pifont}
\usepackage{color}
\usepackage{lineno}
\usepackage{makecell} 
\usepackage{pdflscape}
\usepackage{adjustbox}
\usepackage[utf8]{inputenc}
\usepackage{tabularx}
\usepackage{blindtext}
\usepackage{setspace}
\usepackage{multirow}
\usepackage{notoccite} 
\usepackage{lscape} 
\usepackage{caption} 
\usepackage{mwe}
\usepackage{booktabs}
\usepackage{hyperref}
\usepackage{amsthm}

\newcommand{\RNum}[1]{\lowercase\expandafter{\romannumeral #1\relax}}
\newcommand{\RNumU}[1]{\uppercase\expandafter{\romannumeral #1\relax}}
\usepackage{natbib}
\journal{Elsevier}

\begin{document}
\begin{frontmatter}
\title{Enhancing Robustness and Efficiency of Least Square Twin SVM via Granular Computing}
\author[inst1]{M. Tanveer\texorpdfstring{\corref}{corref}{Correspondingauthor}}\ead{mtanveer@iiti.ac.in}
\author[inst1,inst2]{R. K. Sharma}\ead{rsharma459@gatech.edu}
\author[inst1]{A. Quadir}\ead{mscphd2207141002@iiti.ac.in}
\author[inst1]{M. Sajid }\ead{phd2101241003@iiti.ac.in}

\affiliation[inst1]{organization={Department of Mathematics, Indian Institute of Technology Indore},
            addressline={Simrol}, 
            city={Indore},
            postcode={453552}, 
            state={Madhya Pradesh},
            country={India}}
            \cortext[Correspondingauthor]{Corresponding author}
\affiliation[inst2]{organization={Present Affiliation: H. Milton Stewart School of Industrial and Systems Engineering, Georgia Institute of Technology},
            addressline={}, 
            city={Atlanta},
            postcode={30332}, 
            state={GA},
            country={USA}}
            
\begin{abstract}
In the domain of machine learning, least square twin support vector machine (LSTSVM) stands out as one of the state-of-the-art models. However, LSTSVM suffers from sensitivity to noise and outliers, overlooking the SRM principle and instability in resampling. Moreover, its computational complexity and reliance on matrix inversions hinder the efficient processing of large datasets. As a remedy to the aforementioned challenges, we propose the robust granular ball LSTSVM (GBLSTSVM). GBLSTSVM is trained using granular balls instead of original data points. The core of a granular ball is found at its center, where it encapsulates all the pertinent information of the data points within the ball of specified radius. To improve scalability and efficiency, we further introduce the large-scale GBLSTSVM (LS-GBLSTSVM), which incorporates the SRM principle through regularization terms.  Experiments are performed on UCI, KEEL, and NDC benchmark datasets; both the proposed GBLSTSVM and LS-GBLSTSVM models consistently outperform the baseline models. The source code of the proposed GBLSTSVM  model is available at  \url{https://github.com/mtanveer1/GBLSTSVM}.
\end{abstract}

\end{frontmatter}
\section{Introduction}
Support vector machine (SVM) \cite{cortes1995support} is a powerful model in machine learning that uses kernels to precisely determine the best hyperplane between classes in classification tasks. SVM provides a deterministic classification result. Hence, its application can be found in various domains such as health care \cite{zago2017predicting}, anomaly detection \cite{miao2018distributed}, electroencephalogram (EEG) signal classification \cite{richhariya2018eeg}, Alzheimer’s disease diagnosis \cite{richhariya2020diagnosis}, and so on. SVM implements the structural risk minimization (SRM) principle and, hence, shows better generalization performance. SVM solves a convex quadratic programming problem (QPP) to find the optimal separating hyperplane. However, the effectiveness and efficiency of SVM are limited when dealing with large datasets due to the increase in computational complexity. Further, SVM is sensitive to noise, especially along the decision boundary, and is unstable to resampling.  To mitigate the effects of noise and outliers in data points, fuzzy SVM (FSVM) \cite{lin2002fuzzy, quadir2024intuitionistic} was proposed.  The incorporation of pinball loss in SVM \cite{huang2013support} led to a better classifier that has the same computational complexity as SVM but is insensitive to noise and stable to resampling. To overcome the issue of the computational complexity of SVM, some non-parallel hyperplane-based classifiers have been proposed, such as generalized eigenvalue proximal SVM (GEPSVM) \cite{mangasarian2005multisurface} and twin SVM (TSVM)  \cite{4135685}. The GEPSVM and TSVM generate non-parallel hyperplanes that position themselves closer to the samples of one class while maximizing their distance from samples belonging to the other class. GEPSVM  solves two generalized eigenvalue problems, and its solutions are determined by selecting the eigenvectors corresponding to the smallest eigenvalue. However, TSVM solves two smaller QPPs, which makes its learning process approximately four times faster than SVM \cite{4135685}. Also, TSVM shows better generalization performance than GEPSVM \cite{4135685}. The efficiency of TSVM may decrease due to its vulnerability to noise and outliers, potentially leading to unsatisfactory results. Moreover, its substantially high computational complexity and reliance on matrix inversions pose significant challenges, especially when dealing with large datasets, thereby impeding its real-time applications. Also, TSVM does not adhere to the SRM principle, making the model susceptible to overfitting. \citet{kumar2009least} proposed least squares TSVM (LSTSVM). Unlike TSVM, LSTSVM incorporates an equality constraint in the primal formulation instead of an inequality constraint. This modification allows LSTSVM to train much faster than TSVM, as it solves a system of linear equations to determine the optimal nonparallel separating hyperplanes. However, despite its success in reducing training time, the methodology of LSTSVM involves the computation of matrix inverses, which limits its applicability to large datasets. Additionally, the LSTSVM's ability to learn decision boundaries can be significantly affected by the presence of noisy data and outliers. Some recent advancements in TSVM and LSTSVM include large scale fuzzy LSTSVM for class imbalance learning (LS-FLSTSVM-CIL) \cite{ganaie2022large}, large scale pinball TSVM (LSPTSVM) \cite{tanveer2022large}, smooth linear programming TSVM (SLPTSVM) \cite{tanveer2015application}, 
capped $l_{2, p}$-norm metric-based robust LSTSVM for pattern classification \cite{YUAN2021457}, the Laplacian $l_p$ norm LSTSVM \cite{xie2023laplacian}, sparse solution of least-squares twin multi-class support vector machine using $l_0$ and $l_p$-norm for classification and feature selection \cite{MOOSAEI2023471}, the inverse free reduced universum TSVM for imbalanced data classification \cite{moosaei2023inverse}, and intuitionistic fuzzy weighted least squares TSVM (IFW-LSTSVM) \cite{9794299}. These advancements contribute to the ongoing progress in TSVM and LSTSVM techniques. A comprehensive overview of the various versions of TSVM can be found in \cite{tanveer2022comprehensive}.

The concept of ``large-scale priority" aligns with the natural information-processing mechanism of the human brain \cite{zhou2008neural}. Granulation, or breaking down information into smaller, more manageable parts, is a fundamental aspect of the learning process \cite{hu2022multi}. Our brains are wired to absorb and process information in layers, starting with a broader concept and then delving into the specifics as needed. This approach allows us to grasp the big picture first and then gradually fill in the details, leading to a more comprehensive understanding. Drawing inspiration from the brain's functioning, granular computing explores problems at various levels of detail \cite{wang2023trilevel}. Coarser granularity emphasizes important components, thereby enhancing the effectiveness of learning and resistance to noise. Conversely, finer granularity offers intricate insights that deepen knowledge. In contrast, a majority of machine learning models rely on pixels or data points for training at the lowest possible resolution. Consequently, they are often more susceptible to outliers and noise. This approach lacks the efficiency and scalability of the brain's adaptable granulating capabilities. The novel classifier based on granular computing and SVM \cite{xia2019granular} was developed to incorporate the concept of granular balls. This classifier utilizes hyper-balls to partition datasets into different sizes of granular balls \cite{xia2020fast}. As highlighted in \cite{xia2021granular}, this approach, which imitates cognitive processes observed in the human brain, offers a scalable, dependable, and efficient solution within the realm of granular computing by introducing larger granularity sizes. However, this transition may compromise the accuracy of fine details. Conversely, finer granularity enhances the focus on specific features, potentially improving precision but also introducing challenges in terms of robustness and efficiency in noisy scenarios. Consequently, striking the right balance between granularity and size becomes imperative. 
Researchers persistently explore novel applications, refine existing methodologies \cite{zhang2018multi}, and bridge interdisciplinary gaps to harness the potential of granular computing across diverse domains.

Recently, the granular ball SVM (GBSVM) \cite{xia2022gbsvm} has been proposed, which integrates the concepts of SVM with granular computing. GBSVM addresses a single quadratic programming problem using the PSO algorithm, which can sometimes lead to convergence at local minima. To overcome this limitation, \citet{quadir2024granular} introduced the Granular Ball TSVM (GBTSVM) and its large-scale variant. These models solve two complex quadratic programming problems, improving the performance and robustness of the model. GBTSVM exhibit good performance in effectively managing datasets that are contaminated with noise and outliers. Several other variants of the GB-based classification models have been proposed, such as GBTSVM based on robust loss function \cite{quadir2024granularpin}, enhanced feature-based GBTSVM \cite{quadir2024Enhanced}, and randomized neural networks based on granular computing \cite{sajid2024gb}.
Motivated by the robustness and efficiency demonstrated by the  GBTSVM, we incorporate the concept of granular computing into the LSTSVM and propose a novel model called the granular ball least square twin support vector machine (GBLSTSVM). This integration aims to address the inherent drawbacks and complexities associated with LSTSVM. GBLSTSVM uses granular balls as input to construct non-parallel separating hyperplanes by solving a system of linear equations like in the case of LSTSVM. The granular balls are characterized by their center and radius. The construction of granular balls based on granular computing is elaborated in Section II (B).  The essence of a granular ball lies in its center, which encapsulates all the relevant information of the data points that lie within the ball. In comparision to LSTSVM, GBLSTSVM provides enhanced efficiency, a heightened resistance to noise and outliers, robustness to resampling, and is trained using a substantially reduced number of training instances, thereby significantly reducing the training time. However, GBLSTSVM lacks the SRM principle, which can lead to the potential risk of overfitting. To address this, we further propose the novel large-scale GBLSTSVM (LS-GBLSTSVM) model. LS-GBLSTSVM incorporates the regularization terms in its primal form of the optimization problem which eliminates the need for matrix inversions and also helps to mitigate the risk of overfitting. The main highlights of this paper are as follows:
\begin{enumerate}
        \item We propose the novel GBLSTSVM by incorporating granular computing in LSTSVM. The GLSTSVM is trained by feeding granular balls as input instead of data points for constructing optimal non-parallel separating hyperplanes. The use of granular balls reduces the training time by a substantial amount, amplifies the model's performance, and elevates the robustness against noise and outliers.
        
        \item We propose the novel LS-GBLSTSVM by implementing the SRM principle through the inclusion of regularization terms in the primal formulation of GBLSTSVM. LS-GBLSTSVM does not require matrix inversion, making it suitable for large-scale problems. In addition, LS-GBLSTSVM offers robust overfitting control, noise and outlier resilience, and improved generalization performance.
        
        \item We present the meticulous mathematical frameworks for both GBLSTSVM and LS-GBLSTSVM on linear and Gaussian kernel spaces. The formulation integrates the centers and radii of all granular balls used in training into the LSTSVM model. These models excel in capturing complex data patterns and relationships through sophisticated nonlinear transformations.

         \item We conducted the experiments of our proposed GBLSTSVM and LS-GBLSTSVM models using 34 UCI and KEEL datasets with and without label noise. Our comprehensive statistical analyses demonstrate the significantly superior generalization abilities of our proposed models compared to LSTSVM and the other baseline models. Further, we performed experiments on NDC datasets of sample sizes ranging from 10,000 to 5 million to determine scalability. The results demonstrated that our proposed models surpass the baseline models in terms of accuracy, efficiency, robustness, and scalability.
\end{enumerate}
The subsequent sections of this paper are structured as follows: in Section \ref{relatedworks}, an overview of related work is provided. In Sections III and IV, we present the mathematical framework of our proposed novel GBLSTSVM and LS-GBLSTSVM in both linear and Gaussian kernel spaces, and the computational complexity is discussed in Section V. The experimental results and discussions are presented in Section VI, followed by the conclusions and recommendations for future research are provided in Section \ref{7}.

\section{Related Works}
This section begins with establishing notations and then reviews the concept of granular computing. Also, we briefly outline the mathematical formulations along with their solutions of GBSVM and LSTSVM models. 
\label{relatedworks}
\subsection{Notations}
 Let the training dataset $ T = \bigl\{(x_i,y_i)\vert\, x_i \in \mathbb{R}^{1 \times N},\, y_i \in \{-1,1\}, i = 1, 2, \cdots, m\bigl\}$, where $x_i$  represent the feature vector of each data sample and $y_i$ represents the label of the individual data. Let the number of granular balls generated on $T$ be $\{GB_1, GB_2, \dots, GB_k\}$. Let $o_j$ and $d_j$ be the center and radius of the granular ball $GB_j$. Let  $C \in \mathbb{R}^{k_1 \times N}$ and $ D \in \mathbb{R}^{k_2 \times N}$ be the feature matrices of the centers of the granular balls with label $+1$ class and $-1$ class, respectively, such that $k_1 +k_2 = k$ is the total number of granular balls generated on $T$. Let $R^{+} \in \mathbb{R}^{k_1 \times 1} $ and $R^{-} \in \mathbb{R}^{k_2 \times 1} $ be the column vectors containing the radius of all granular balls having label $+1$ and $-1$, respectively. $({\cdot})^{'}$ is the transpose operator and $e_i$ represents the column vectors of ones of appropriate dimensions.
 
\subsection{Granular Computing} 
In 1996, Lin and Zadeh proposed the concept of ``granular computing''. It becomes computationally expensive to process every data point in the space when dealing with large datasets. The objective of granular computing is to reduce the number of training data points required for machine learning models. The core idea behind granular computing is to use granular balls to completely or partially cover the sample space. This captures the spirit of data simplification while maintaining representativeness during the learning process. Granular balls, characterized by two parametric simple representations, a center $o$ and a radius $d$,  are the most appropriate choice for effectively handling high-dimensional data. Given a granular ball $(GB)$ containing the datapoints  $\{x_1,x_2, \dots, x_p\}$, where $x_i \in$  $ \mathbb{R}^{1\times N} $, the center $o$ of a $GB$ is the center of gravity for all sample points in the ball, and $d$ is equal to the average distance from $o$ to all other points in $GB$. Mathematically, they can be calculated as: $o = \frac{1}{p} \sum_{i=1}^{p}x_i$ and $ d = \frac{1}{p} \sum_{i=1}^{p}||x_i - o||$.
The average distance is utilized to calculate the radius $d$ as it remains unaffected by outliers and aligns appropriately with the distribution of the data. The label assigned to a granular ball is determined by the labels of the data points that have the maximum frequency within the ball. To quantitatively assess the amount of splitting within a granular ball, the concept of ``threshold purity" is introduced. This threshold purity represents the percentage of the majority of samples within the granular ball that possess the same label. The number of granular balls generated on $T$ is given by the following optimization problem:
    \begin{align}
    \text{min}\hspace{0.2cm}&\gamma_1 \times \frac{m}{\sum_{j=1}^{k}\lvert GB_j\rvert} + \gamma_2*k ,\nonumber\\
        \text { s.t. } \hspace{0.1cm} & quality(GB_j) \geq \rho,
    \end{align}
where $\gamma_1$ and $\gamma_2$ are weight coefficients. $\rho$ is the threshold purity. $|.|$ represents the cardinality of a granular ball, and $m$ and $k$ represent the number of samples in  $T$ and the number of granular balls generated on $T$, respectively. 
\begin{figure}[ht!]
 \label{fig:pictorial_rep_GB}
      \centering
       \includegraphics[width=12cm,height=10cm]{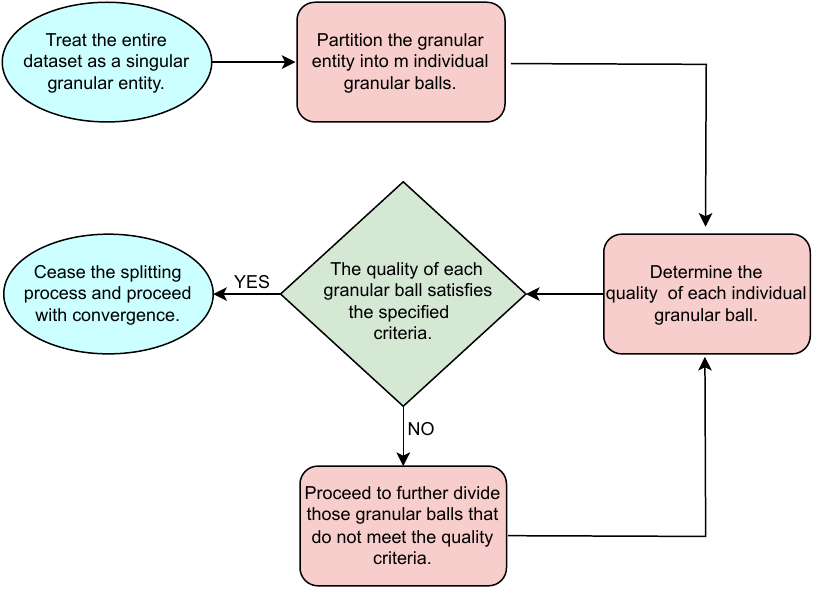}
       \begin{center}
            \caption{Pictorial representation of the process of granular ball generation.}
       \end{center}
      \end{figure}
The quality of each granular ball is adaptive \cite{xia2022efficient}. Initially, the whole dataset is considered as a single granular ball, which fails to accurately represent the dataset's distribution and exhibits the lowest level of purity. In the cases where the purity of this granular ball falls below the specified threshold, it is necessary to divide it multiple times until all sub-granular balls achieve a purity level equal to or higher than the threshold purity. As the purity of the granular balls increases, their alignment with the original dataset's data distribution improves. Fig. 1 depicts the procedure of granular ball generation.

 
\subsection{Least Square Twin Support Vector Machine (LSTSVM)}
Suppose matrix $A$ and $B$ contain all the training data points belonging to the $+1$ and $-1$ class, respectively. The primal problem of LSTSVM \cite{kumar2009least} can be expressed as:
\begin{align}{}
\underset{w_1, b_1}{\text{min}}\hspace{0.2cm}&\frac{1}{2}||Aw_1  + e_1b_1||^2+ \frac{c_1}{2}||q_1||^2\nonumber,\\
\text { s.t. } \hspace{0.2cm}& -(Bw_1 + e_2b_1) + q_1 = e_2,
\end{align} 
\hspace{1cm}and 
\begin{align}
  \underset{w_2,b_2}{\text{min}}\hspace{0.2 cm}   &\frac{1}{2}||Bw_2 +e_2b_2||^2 + \frac{c_2}{2} ||q_2||^2 ,\nonumber \\
     \text{s.t.} \hspace{0.2 cm} &(Aw_2 + e_1b_2) + q_2 = e_1 ,
\end{align}
where $q_1$ and $q_2$ are slack variables. 
Substituting the equality constraint into the primal problem, we get
\begin{align}
     \underset{w_1 , b_1}{\text{min}} \hspace{0.2 cm} \frac{1}{2}||Aw_1 + e_1b_1||^2 + \frac{c_1}{2}||Bw_1 +e_2b_1 +e_2||^2,
\end{align} 
and 
\begin{align}
     \underset{w_2 , b_2}{\text{min}}\hspace{0.2 cm} \frac{1}{2}||Bw_2 + e_2b_2||^2 + \frac{c_2}{2}||Aw_2 +e_1b_2 - e_1||^2.
\end{align} 
Taking gradient of (4) with respect to $w_1$ and $b_1$ and solving, we get
\begin{equation}
    \begin{bmatrix}
    w_1\\
    b_1
\end{bmatrix} 
= -
\begin{bmatrix}
    F'F + \frac{1}{c_1}E'E
\end{bmatrix}^{-1}F'e_2 .
\end{equation}
Similarly,
\begin{equation}
    \begin{bmatrix}
    w_2\\
    b_2
\end{bmatrix} 
= 
\begin{bmatrix}
    E'E + \frac{1}{c_2}F'F
\end{bmatrix}^{-1}E'e_1 , 
\end{equation}
where $E=\begin{bmatrix}
   A & e_1
\end{bmatrix}$ and $F =\begin{bmatrix}
    B & e_2
\end{bmatrix}.$\\

Once the optimal values of $w_1, b_1$ and $w_2, b_2$ are calculated. The
categorization of a new input data point x  $\in \mathbb{R}^{1 \times N}$ into either the $+1$
or $-1$ class can be determined as follows:
\begin{align}{}
\text{Class(x)} =  \underset{i \in \lbrace 1,2 \rbrace}{\text{argmin}} \Big(\frac{\|w_i\text{x}+b_i\|}{\|w_i\|}\Big).
\end{align} 

\subsection{ Granular Ball Support Vector Machine (GBSVM)} 
The basic idea of GBSVM \cite{xia2022gbsvm} is to mimic the classical SVM \cite{cortes1995support}  using granular balls during the training process instead of data points. This makes GBSVM efficient and robust in comparison to SVM. The parallel hyperplanes in GBSVM are constructed using supporting granular balls $GB_j$ having support center $o_j$ and support radius $d_j$. Fig. 2 depicts the construction of inseparable GBSVM using supporting granular balls.
 \begin{figure}[ht!]
       \centering
        \includegraphics[width=9cm, height=7cm]{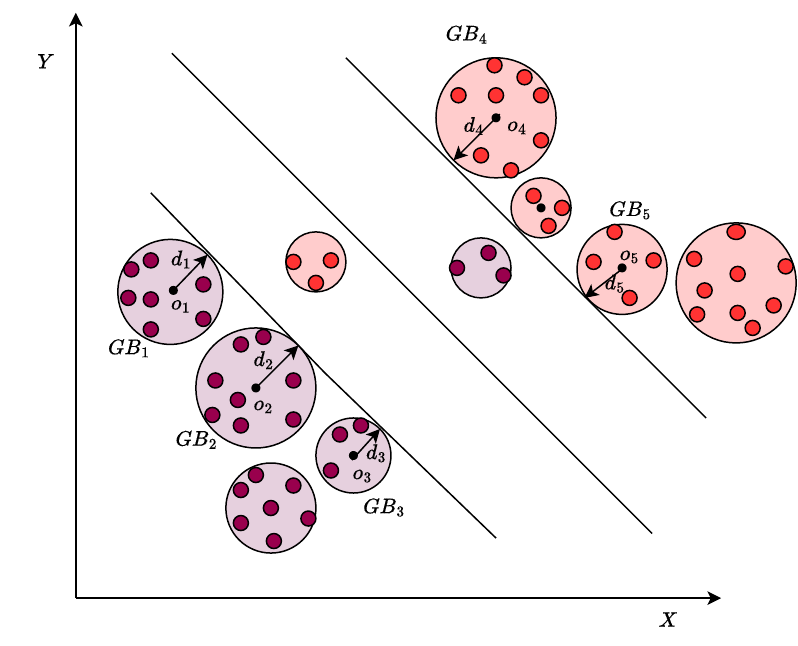}
        \caption{Inseperable GBSVM model having supporting granular  balls $GB_j$, supporting centers $o_j$, and supporting radii $d_j$. }
        \label{fig:enter-label}
\end{figure}
\\
The inseparable GBSVM model can be expressed as:
\begin{align}
    \underset{w,b,\xi_j}{\text{min}} &\hspace{0.4cm} \frac{1}{2}\|w\|^2+ C\sum_{j=1}^{k} \xi_j, \nonumber\\
    \text{s.t.}&\hspace{0.5cm} y_j(wo_j +b)- \|w\|d_j \geq 1 - \xi_j, \nonumber \\
    &\hspace{0.5cm}\xi_j \geq 0, \hspace{0.2cm}j = 1,2,3,...,k.
\end{align}
The dual of inseparable GBSVM formulation is:
\begin{align}
\underset{\alpha}{\text{max}}&\hspace{0.2cm} -\frac{1}{2}\|w\|^2 + \sum_{j=1}^{k} \alpha_j, \nonumber \\
\text{s.t.}&\hspace{0.2cm}\sum_{j=1}^{k}\alpha_jy_j= 0,\nonumber \\ 
 &0\leq \alpha_j \leq C,  \hspace{0.2cm}j = 1,2,3,...,k, 
\end{align}
where $\alpha_j$'s are Lagrange multipliers.

\section{THE PROPOSED GRANULAR BALL LEAST SQUARE TWIN SUPPORT VECTOR MACHINE (GBLSTSVM)}
In this section, we introduce a novel GBLSTSVM to tackle the binary classification problem. We propose the use of granular balls that encompass either the complete sample space or a fraction of it during the training process. These granular balls, derived from the training dataset, are coarse and represent only a small fraction of the total training data points. This coarse nature renders our proposed model less susceptible to noise and outliers. 

By leveraging the granular balls, we aim to generate separating hyperplanes that are nonparallel and can effectively classify the original data points. In the construction of optimal separating hyperplanes, we aim to utilize the maximum information stored in all the training data points while simultaneously decreasing the data points required to find optimal separating hyperplanes. 
\begin{figure}[ht!]
    \centering
    \includegraphics[width=11.00cm, height=9cm]{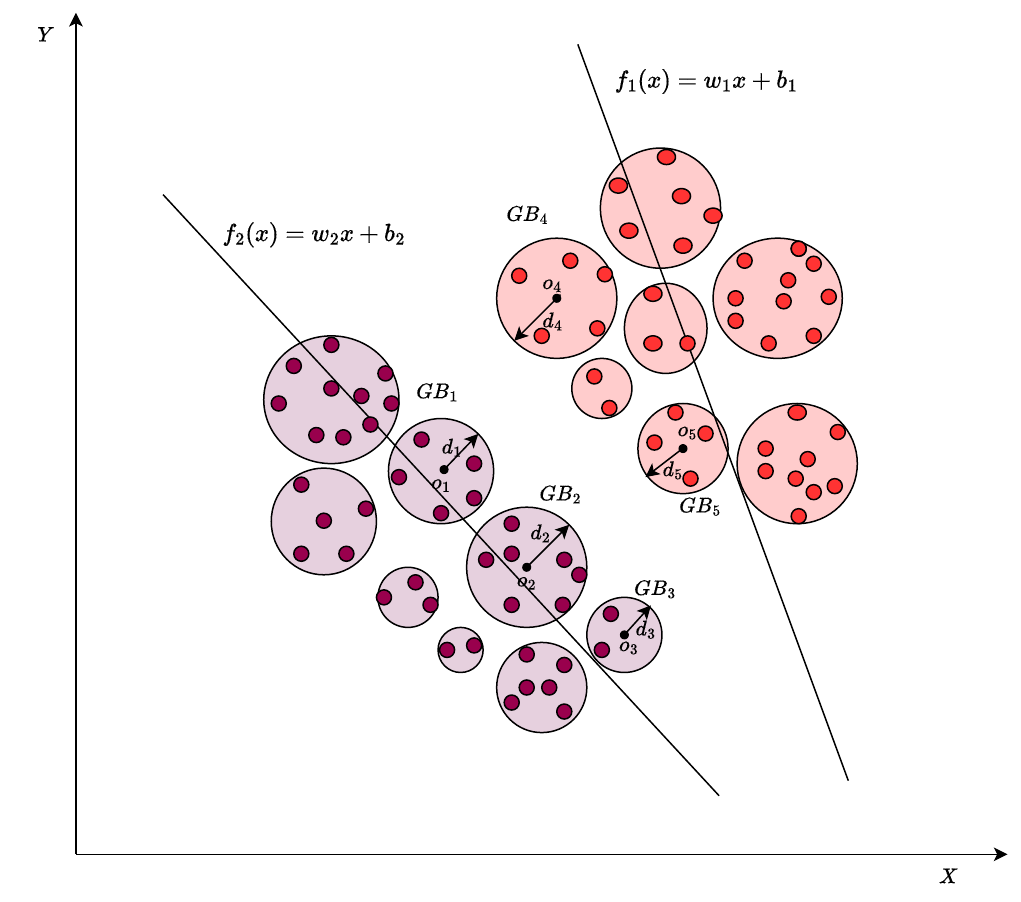}
    \caption{Granular ball least square twin support vector machine model having two non-parallel hyperplanes $f_1$ and $f_2$. }
    \label{fig:enter-label1}
\end{figure}

Hence, we incorporate both the centers and radii of all the granular balls generated through granular computing into the primal formulation of LSTSVM. The integration of granular balls in the training process not only enhances the LSTSVM's robustness against noise and outliers but also significantly reduces the training time, offering enhanced robustness and reduced computational complexity. The geometric representation of the GBLSTSVM model is illustrated in Figure 3. 
\subsection{Linear GBLSTSVM:}
The optimized formulation of linear GBLSTSVM is given by: 
\begin{align}{}
\label{eq:12}
\underset{w_1,  b_1 , q_1}{\text{min}}\hspace{0.2cm}&\frac{1}{2}\|Cw_1  + e_1b_1\|^2 + \frac{c_1}{2}\|q_1\|^2\nonumber\\
\text {s.t.} \hspace{0.2cm}& -(Dw_1 + e_2b_1) + q_1 = e_2 + R^- 
\end{align}
\hspace{1cm}and
\begin{align}
\label{eq:13}
\underset{w_2,b_2, q_2}{\text{min}}\hspace{0.2 cm}&\frac{1}{2}\|Dw_2 +e_2b_2\|^2 + \frac{c_2}{2}\|q_2\|^2 \nonumber \\
\text{s.t.}\hspace{0.2 cm}&(Cw_2 + e_1b_2) + q_2 = e_1 + R^+,   
\end{align}
where $q_1$ and $q_2$ are slack variables and  $c_1$ and $c_2$ are tunable parameters. $\begin{bmatrix} w_1 ;& b_1 \end{bmatrix}$ and $\begin{bmatrix} w_2; & b_2 \end{bmatrix}$ are the hyperplane parameters. 
 In the above equations (\ref{eq:12}) and (\ref{eq:13}), the matrices \( C \in \mathbb{R}^{k_1 \times N}\) and \( D \in \mathbb{R}^{k_2 \times N}\) represent the centers, while the vectors \( R^+ \in \mathbb{R}^{k_1 \times 1} \) and \( R^- \in \mathbb{R}^{k_2 \times 1}\) denote the radii of the granular balls corresponding to the class labels \( +1 \) and \( -1 \), respectively. The optimization problems are designed to incorporate the maximum information from the training dataset by leveraging the structural representations provided by \( C \) and \( D \), and the radii \( R^+ \) and \( R^- \), which collectively capture the distribution and scale of the data for each class. The centers (\( C \) and \( D \)) summarize the spatial distribution of the data clusters, allowing the optimization to focus on the overall structure rather than individual data points. The radii (\( R^+ \) and \( R^- \)) quantify the spread or variability of the clusters within each class, making the model robust to noise and outliers. This approach shifts the emphasis from individual outlier data points to the broader class distribution, ensuring stability and generalization. The least squares formulation minimizes the squared error between the hyperplane and the centers of the granular balls:
\[
\|Cw_1 + e_1b_1\|^2 \quad \text{and} \quad \|Dw_2 + e_2b_2\|^2,
\]
where \( w_1, b_1 \) and \( w_2, b_2 \) are the parameters of the hyperplanes. By minimizing these errors, the hyperplanes align closely with the granular ball centers, ensuring the model effectively utilizes the structural information inherent in the data. The constraints incorporate the radii (\( R^+ \) and \( R^- \)) to account for the scale and variability of the data. This ensures that the separation between classes is proportional to their intrinsic distributions, enabling the model to capture meaningful patterns and relationships within the data while maintaining robustness and scalability.


To solve \eqref{eq:12}, we substitute the equality constraint into the primal problem
\begin{align*}
     \underset{w_1 , b_1}{\text{min}} \hspace{0.2 cm} \frac{1}{2}\|Cw_1 + e_1b_1\|^2 + \frac{c_1}{2}\|Dw_1 +e_2b_1 +e_2 + R^-\|^2.
\end{align*}
\\
   Taking gradient with respect to $w_1$ and $b_1$ and equating to 0, we get 
   $$ C'(Cw_1 +e_1b_1) + c_1D'(Dw_1 +e_2b_1 +e_2 + R^-)=0$$
   and
  $$ e_1'(Cw_1 +e_1b_1) + c_1e_2'(Dw_1 +e_2b_1 +e_2 + R^-)=0.$$
Converting the system of linear equations into matrix form and solving for $w_1$ and $b_1$ we get
\begin{align*}
\begin{bmatrix}
     D'D  & D'e_2 \\
     e_2'D & m_2
\end{bmatrix}
\begin{bmatrix}
    w_1\\
    b_1\end{bmatrix}
 + 
 \frac{1}{c_1}
\begin{bmatrix}
    C'C & C'e_1 \\
    e_1'C  & m_1
\end{bmatrix}
\begin{bmatrix}
    w_1\\
    b_1
\end{bmatrix}  
+ 
\begin{bmatrix}
    D'e_2 + D'R^-\\
    m_2 + e_2'R^-
\end{bmatrix} 
= 0,
\end{align*}
where $e_1'e_1 = m_1$ and $e_2'e_2 = m_2$.
\begin{align} \implies
\begin{bmatrix}
    w_1\\
    b_1 \end{bmatrix} = &- \begin{bmatrix}
    D'D  +  \frac{1}{c_1}C'C & D'e_2 +  \frac{1}{c_1}C'e_1 \\
     e_2'D  +  \frac{1}{c_1}e_1'C & m_2  + \frac{1}{c_1}m_1
\end{bmatrix}^{-1}   \begin{bmatrix}
    D'e_2 + D'R^-\\
    m_2 + e_2'R^- \end{bmatrix},\\ \nonumber
    \implies
\begin{bmatrix}
    w_1\\
    b_1  \end{bmatrix} = &- \begin{bmatrix}
    \begin{bmatrix}
   D' \\
   e_2'
\end{bmatrix}
\begin{bmatrix}
    D & e_2
\end{bmatrix}
+ 
\frac{1}{c_1}
\begin{bmatrix}
    C' \\
    e_1'
\end{bmatrix}
\begin{bmatrix}
   C & e_1
\end{bmatrix}
\end{bmatrix}^{-1} \begin{bmatrix}
    D'& D'\\
    e_2' & e_2'
\end{bmatrix}
\begin{bmatrix}
    e_2\\
    R^-
\end{bmatrix},\\
\implies
\begin{bmatrix}
    w_1\\
    b_1 \end{bmatrix} 
 = & -
\begin{bmatrix}
    F'F + \frac{1}{c_1}E'E
\end{bmatrix}^{-1}\overline{F} \overline{e_2},
\end{align}

where \begin{align} E= &\begin{bmatrix}
   C & e_1
\end{bmatrix},\hspace{0.25cm} F =\begin{bmatrix}D & e_2 \end{bmatrix},\hspace{0.25cm}  \overline{F} =&\begin{bmatrix} D' & D' \\ {e_2}' & {e_2}' \end{bmatrix}, \text{and} \hspace{0.25cm} \overline{e_2} = \begin{bmatrix}e_2\\ R^-\end{bmatrix}. 
\end{align}
\\
Solving \eqref{eq:13} in a similar way, we get

\begin{align}
    \begin{bmatrix}
    w_2\\
    b_2
\end{bmatrix} 
& = 
\begin{bmatrix}
    E'E + \frac{1}{c_2}F'F
\end{bmatrix}^{-1}
\begin{bmatrix}
    C'e_1 + C'R^+\\
    m_1 + e_1'R^+
\end{bmatrix},\\
\begin{bmatrix}
    w_2\\
    b_2
\end{bmatrix} 
& = 
\begin{bmatrix}
    E'E + \frac{1}{c_2}F'F
\end{bmatrix}^{-1}
\overline{E} \overline{e_1},
\end{align}
where  $\overline{E} =\begin{bmatrix} C'& C'\\e_1' & e_1'\end{bmatrix}$ and $\overline{e_1}= \begin{bmatrix}e_1\\R^+ \end{bmatrix}$.

Once the optimal values of $w_1, b_1$ and $w_2, b_2$ are calculated. The
categorization of a new input data point x  $\in \mathbb{R}^{1 \times N}$ into either the $+1$ or $-1$ class can be determined as follows:
\begin{align}{}
\text{Class(x)} =  \underset{i \in \lbrace 1,2 \rbrace}{\text{argmin}} \Big(\frac{\|w_i\text{x}+b_i\|}{\|w_i\|}\Big).
\end{align}

\begin{figure}[ht!]
    \label{framework}
    \centering
    \includegraphics[width=16.00cm, height=10cm]{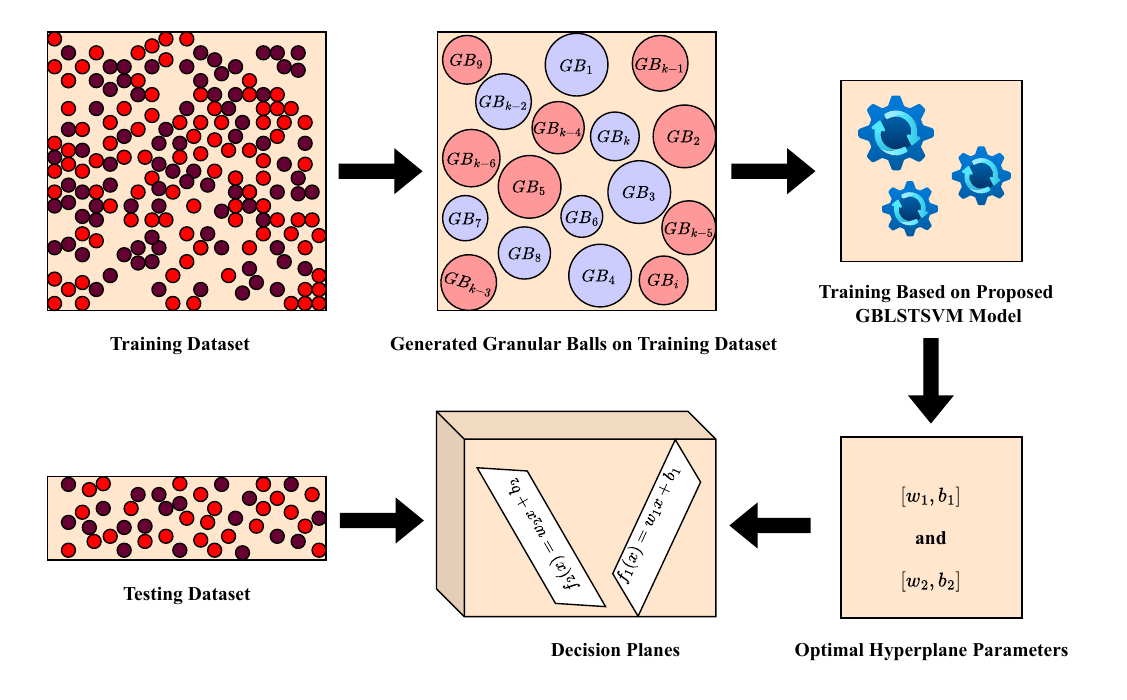}
    \caption{Overall architectural framework of the proposed linear GBLSTSVM model.}
    \label{fig:enter-label2}
\end{figure}

The overall architectural framework of the proposed approach is illustrated in Figure 4, depicting its key components and workflow:
\begin{enumerate}
\item Granular ball representation: The training data points are transformed into granular balls using granular computing, which captures the underlying structural distribution of the dataset while reducing sensitivity to noise and outliers.
    \item Hyperparameter optimization: The hyperparameters, $num$, $pur$, $c_1$, and $c_2$, are systematically optimized during training to ensure an adaptive, data-driven learning process that enhances generalization.
    \item Hyperplane parameter estimation: The optimized hyperparameters are then utilized to compute the optimal hyperplane parameters $[w_1; b_1]$ and $[w_2; b_2]$, which define the non-parallel decision boundaries tailored to the granular ball representation.
    \item Classification decision boundaries: The derived hyperplanes are employed to construct robust, non-parallel decision planes, effectively classifying training data points while maintaining high resilience against noise and data variations.
\end{enumerate}

\subsection{Nonlinear GBLSTSVM}
To generalize our proposed model to the nonlinear case, we introduce the map $x^{\phi} = \phi(x): \mathbb{R}^N \to \mathbb{H}$, where $\mathbb{H}$ represents a Hilbert space. We define ${T}^\phi = \lbrace x^\phi : x \in T \rbrace$, where $T$ denotes the training dataset. The granular balls that are generated on the set $T^\phi$ are denoted by $ \mathbb{G}^\phi = \lbrace (({o_1}^\phi, {d_1}^\phi),y_1), (({o_2}^\phi, {d_2}^\phi),y_2), \\ \cdots, (({o_k}^\phi, {d_k}^\phi),y_k) \rbrace $, where $k$ represents the number of granular balls. Let the matrices $C^\phi$ and $D^\phi$, along with column vectors ${R_\phi}^+$ and ${R_\phi}^-$, represent the features of the centers and radii of the granular balls belonging to the positive and negative class, respectively.

The optimization problem for nonlinear GBLSTSVM is given as:
\begin{align}{} 
\label{eqn:21}
\underset{w_1, b_1 , q_1}{\text{min}}\hspace{0.2cm}&\frac{1}{2}\|C^\phi w_1  + e_1b_1\|^2 + \frac{c_1}{2}\|q_1\|^2,\nonumber\\
\text { s.t. } \hspace{0.2cm}& -(D^\phi w_1 + e_2b_1) + q_1 = e_2 + {R_\phi}^-, 
\end{align}
\hspace{1cm}and
 \begin{align}
 \label{eqn:22}
\underset{w_2, b_2 , q_2}{\text{min}}\hspace{0.2cm}&\frac{1}{2} \|D^\phi w_2  + e_2b_2\|^2 + \frac{c_2}{2}\|q_2\|^2,\nonumber\\
\text { s.t. } \hspace{0.2cm}& (C^\phi w_2 + e_1b_2) + q_2 = e_1 + {R_\phi}^+.
\end{align}
The solutions of \eqref{eqn:21} and \eqref{eqn:22} can be derived similarly as in the linear case. The solutions are:
\begin{align}
    \begin{bmatrix}
    w_1\\
    b_1
\end{bmatrix} 
&= -
\begin{bmatrix}
    H'H + \frac{1}{c_1}G'G
\end{bmatrix}^{-1}\overline{H} \overline{e_2},
\end{align}
\hspace{1cm}and
\begin{align}
\begin{bmatrix}
    w_2\\
    b_2
\end{bmatrix} 
&= 
\begin{bmatrix}
    G'G+ \frac{1}{c_2}H'H
\end{bmatrix}^{-1}
\overline{G} \overline{e_1},
\end{align}
where   
\begin{align}
H &=\begin{bmatrix} D^\phi & e_2 \end{bmatrix}, \hspace{1cm} G = \begin{bmatrix} C^\phi & e_1\end{bmatrix},
\\
\overline{H} &=\begin{bmatrix} {D^\phi}' & {D^\phi}' \\{e_2}' & {e_2}' \end{bmatrix},  \hspace{0.6cm} \overline{G} =\begin{bmatrix}  {C^\phi}'&  {C^\phi}'\\{e_1}' & {e_1}'\end{bmatrix} ,\\ \overline{e_1}& =\begin{bmatrix}e_1\\{R_\phi}^+\end{bmatrix},\hspace{1.4cm}\overline{e_2}= \begin{bmatrix}e_2\\{R_\phi}^- \end{bmatrix}.
 \end{align}
The classification of data points to class $ +1 $ or $-1$ is done similarly to the linear case of the GBLSTSVM model. 

\section{THE PROPOSED LARGE SCALE GRANULAR BALL LEAST SQUARE TWIN SUPPORT VECTOR MACHINE (LS-GBLSTSVM)}
Granular computing significantly reduces the number of training instances, leading to a substantial reduction in computational requirements. However, the scalability of the GBLSTSVM may decrease when confronted with large datasets due to its reliance on matrix inversion for solving the system of linear equations. Additionally, like LSTSVM, GBLSTSVM lacks the SRM principle. To address these issues, we introduce a regularization term into the primal formulation of GBLSTSVM. This inclusion results in an additional equality constraint in the primal formulation, effectively eliminating the necessity for matrix inversions in obtaining optimal nonparallel hyperplanes in GBLSTSVM. This removal of matrix inversions significantly reduces the computational complexity of LS-GBLSTSVM, making it well-suited for handling large datasets. Moreover, the integration of the regularization terms implements the SRM principle in GBLSTSVM.
\subsection{Linear LS-GBLSTSVM}
The optimized primal problem of linear LS-GBLSTSVM is given by:
\begin{align}{}
\label{eq:28}
\underset{w_1, b_1 , q_1 ,\eta_1}{\text{min}}&\hspace{0.2cm}\frac{c_3}{2}(\|w_1\|^2  + b_1^2) + \frac{1}{2}\|\eta_1\|^2 + \frac{c_1}{2}\|q_1\|^2,\nonumber\\
\text { s.t. }& \hspace{0.2cm}\eta_1 = Cw_1 + e_1b_1, \nonumber\\
\hspace{0.2cm}& -(Dw_1 + e_2b_1) + q_1 = e_2 + R^-,
\end{align}
\hspace{1cm}and
\begin{align}{}
\label{eq:29}
\underset{w_2, b_2 , q_2 ,\eta_2}{\text{min}}\hspace{0.2cm}&\frac{c_4}{2}(\|w_2\|^2  + b_2^2) + \frac{1}{2}\|\eta_2\|^2 + \frac{c_2}{2}\|q_2\|^2,\nonumber\\
\text { s.t. } \hspace{0.2cm} & \eta_2 = Dw_2 + e_2b_2, \nonumber \\
\hspace{0.2cm}& (Cw_2 + e_1b_2) + q_2 = e_1 + R^+.
\end{align}
Introducing Lagrange multipliers $\alpha \in \mathbb{R}^{k_1\times 1}$ and $\beta \in \mathbb{R}^{k_2\times 1}$ in \eqref{eq:28}, we get
\begin{align}
\label{eq:30}
    L = &\frac{c_3}{2}(\|w_1\|^2  + b_1^2) + 
    \frac{1}{2}\|\eta_1\|^2 + \frac{c_1}{2}\|q_1\|^2  \nonumber \\
    & + \alpha'(\eta_1 - Cw_1 - e_1b_1)\nonumber \\ 
    &+ \beta'(-(Dw_1 + e_2b_1) -e_2 - R^- + q_1).
\end{align}
Applying the K.K.T. necessary and sufficient conditions for \eqref{eq:30} we obtain the following:
\begin{align}
\label{eq:31}
\frac{\partial L}{\partial w_1} &= c_3w_1 - C'\alpha - D'\beta = 0, \\
\label{eq:32}
\frac{\partial L}{\partial b_1} &= c_3b_1 - e_1'\alpha - e_2'\beta = 0, \\ 
\label{eq:33}
\frac{\partial L}{\partial q_1} &= c_1q_1 + e_2'\beta = 0, \\
\label{eq:34}
\frac{\partial L}{\partial \eta_1} &= \eta_1 + \alpha = 0,\\
\label{eq:35}
 \eta_1 &= Cw_1 + e_1b_1, 
\end{align}
\begin{equation}
\label{eq:36}
    -(Dw_1 + e_2b_1) + q_1 = e_2 + R^-. 
\end{equation}
From \eqref{eq:31} and \eqref{eq:32}, we get
\begin{equation}
\label{eq:37}
    \begin{bmatrix}
        w_1 \\ b_1
    \end{bmatrix}
    = \frac{1}{c_3}
    \begin{bmatrix}
        C' & D' \\
        e_1' & e_2'
    \end{bmatrix}
    \begin{bmatrix}
        \alpha \\ \beta
    \end{bmatrix}.
\end{equation}
Substituting \eqref{eq:33}, \eqref{eq:34}, and \eqref{eq:37} in \eqref{eq:30} and simplifying, we get the dual of \eqref{eq:28}:
\begin{align}
\label{eq:38}
    \underset{\alpha , \beta}{
    \text{max}} & - \frac{1}{2} \begin{pmatrix}
        \alpha' & \beta'
    \end{pmatrix}Q_1\begin{pmatrix}
        \alpha' & \beta'
    \end{pmatrix}'- c_3\beta'(e_2 + R^-)  \nonumber\\
\text{where} \hspace{0.2cm}  &Q_1 = \begin{bmatrix}
    CC' + c_3I_1 & CD' \\
    DC' & DD'+ \frac{c_3}{c_1}I_2
\end{bmatrix} + E.
\end{align}
Here, $E$ is the matrix of ones of appropriate dimensions, and $I_1$ and $I_2$ are the identity matrices.
Similarly, the Wolfe Dual of \eqref{eq:29} is:
\begin{align}
\label{eq;39}
    &\underset{\lambda , \theta}{\text{max}} - \frac{1}{2} \begin{pmatrix}
        \lambda' & \theta'
    \end{pmatrix}Q_2\begin{pmatrix}
        \lambda' & \theta'
    \end{pmatrix}'- c_4\theta'(e_1 + R^+), \nonumber  \\
    \text{where} \hspace{0.2cm}&Q_2 = \begin{bmatrix}
    DD' + c_4I_2 & DC' \\
    CD' & CC'+ \frac{c_4}{c_2}I_1
\end{bmatrix} + E.
\end{align}
Then $w_2, b_2$ is given by:
\begin{equation}
    \begin{bmatrix}
        w_2 \\ b_2
    \end{bmatrix}
    = - \frac{1}{c_4}
    \begin{bmatrix}
        D' & C' \\
        e_2' & e_1'
    \end{bmatrix}
    \begin{bmatrix}
        \lambda \\ \theta
    \end{bmatrix}.
\end{equation}

The categorization of a new input data point x  $\in \mathbb{R}^{1 \times N}$ into either the $+1$ or $-1$ class can be determined as follows:
\begin{align}{}
\text{Class(x)} =  \underset{i \in \lbrace 1,2 \rbrace}{\text{argmin}} \Big(\frac{\|w_i\text{x}+b_i\|}{\|w_i\|}\Big).
\end{align}
\subsection{Nonlinear LS-GBLSTSVM}
The optimization problem of nonlinear LS-GBLSTSVM is given as follows:
\begin{align}{} 
\label{eq:48}
\underset{w_1, b_1 , q_1 ,\eta_1}{\text{min}}\hspace{0.2cm}&\frac{c_3}{2}(\|w_1\|^2  + b_1^2) + \frac{1}{2}\|\eta_1\|^2 + \frac{c_1}{2}\|q_1\|^2, \nonumber\\
\text { s.t. } \hspace{0.2cm} & \eta_1 = C^{\phi}w_1 + e_1b_1, \nonumber\\
\hspace{0.2cm}& -(D^{\phi}w_1 + e_2b_1) + q_1 = e_2 + {R_\phi}^-,
\end{align}
\hspace{1cm}and 
 \begin{align}{} 
 \label{eq:49}
\underset{w_2, b_2 , q_2 ,\eta_2}{\text{min}}\hspace{0.2cm}&\frac{c_4}{2}(\|w_2\|^2  + b_2^2) + \frac{1}{2}\|\eta_2\|^2 + \frac{c_2}{2}\|q_2\|^2,\nonumber\\
\text { s.t. } \hspace{0.2cm} & \eta_2 = D^{\phi}w_2 + e_2b_2, \nonumber\\
\hspace{0.2cm}& (C^{\phi}w_2 + e_1b_2) + q_2 = e_1 + {R_\phi}^+. 
\end{align}
Calculating Lagrangian as in the Linear LS-GBLSTSVM, we get
\begin{equation}
    \begin{bmatrix}
        w_1 \\ b_1
    \end{bmatrix}
    = \frac{1}{c_3}
    \begin{bmatrix}
        {C^\phi}' & {D^\phi}' \\
        {e_1}' & {e_2}'
    \end{bmatrix}
    \begin{bmatrix}
        \alpha \\ \beta
    \end{bmatrix}
\end{equation}
The dual of the optimization problem \eqref{eq:48} and \eqref{eq:49} are, 
\begin{align}
    \underset{\alpha,\beta}{\text{max}} &- \frac{1}{2} \begin{pmatrix}
        \alpha' & \beta' \end{pmatrix}Q_1\begin{pmatrix}
        \alpha' & \beta' \end{pmatrix}'- c_3\beta'(e_2 + {R_\phi}^-),\nonumber\\
\text{where,} \hspace{0.2cm} & Q_1 = \begin{bmatrix}
    C^\phi{C^\phi}' + c_3I_1 & C^\phi{D^\phi}' \\
    D^\phi{C^\phi}' & D^\phi{D^\phi}'+ \frac{c_3}{c_1}I_2
\end{bmatrix} + E,
\end{align}
\hspace{1cm}and
\begin{align}
    \underset{\lambda , \theta}{\text{max}} &-  \frac{1}{2} \begin{pmatrix}
        \lambda' & \theta'
    \end{pmatrix}Q_2\begin{pmatrix}
        \lambda' & \theta'
    \end{pmatrix}'- c_4\theta'(e_1 + {R_\phi}^+),\nonumber \\
\text{where,} \hspace{0.2cm} &Q_2 = \begin{bmatrix}
    D^\phi{D^\phi}' + c_4I_2 & D^\phi{C^\phi}'   \\
    C^\phi{D^\phi}' & C^\phi{C^\phi}'+ \frac{c_4}{c_2}I_1
\end{bmatrix} + E . 
\end{align}
Then $w_2, b_2$ is given by:
\begin{equation}
    \begin{bmatrix}
        w_2 \\ b_2
    \end{bmatrix}
    = - \frac{1}{c_4}
    \begin{bmatrix}
        {D^\phi}' & {C^\phi}' \\
        {e_2}' & {e_1}'
    \end{bmatrix}
    \begin{bmatrix}
        \lambda \\ \theta
    \end{bmatrix}
\end{equation}
To solve the optimization problem of type (38) and (44), we use Sequential Minimal Optimization (SMO)  \cite{keerthi2003smo}. The classification of test data points to class $+1$ or $-1$ is done in the same manner as in the linear case of the LS-GBLSTSVM model. 
\section{Computational Complexity}
\begin{algorithm}
\caption{Linear GBLSTSVM and LS-GBLSTSVM Model Algorithm}
\label{alg:granular_ball_svm}
\begin{algorithmic}[1]
\State Initialize $GB$ as the entire dataset $T$ and set  $G$ as an empty collection: $GB=T$, $G=\{\}$.
\State Initialize $Object$ as a collection containing $GB$: $Object = \{GB\}$.
\For{$j = 1$ to $\lvert Object \rvert$} 
    \If{$pur(GB_j) < \rho$} 
        \State Split $GB_j$ into $GB_{j1}$ and $GB_{j2}$ using 2-means clustering.
        \State Update $Object$ with the newly formed granular balls: $Object \leftarrow GB_{j1}, GB_{j2}$.
    \Else
        \State Calculate center $o_j$ and radius $d_j$ of $GB_j$: 
        \State $o_j = \frac{1}{n_j} \sum_{i=1}^{n_j} x_i$, where $x_i \in GB_j$ and $n_j$ is the number of samples in $GB_j$.
        \State $d_j = \frac{1}{n_j}\sum_{i=1}^{n_j}||x_i - o_j ||$.
        \State Assign label $y_j$ to $GB_j$ based on the majority class samples within $GB_j$.
        \State Add $GB_j = \{((o_j, d_j),y_j)\}$ to $G$.
    \EndIf
\EndFor
\If{$Object \neq \{\}$} 
    \State Repeat steps 3-14 for further splitting.
\EndIf
\State Generate granular balls: $G=\{((o_j, d_j), y_j),\hspace{0.2cm} j=1,2, \ldots, k\}$, where $k$ is the number of granular balls.
\State For GBLSTSVM, compute $w_1$, $b_1$, $w_2$, and $b_2$ using (14) and (17) and for LS-GBLSTSVM, solve (36) and (37) to obtain $\alpha$, $\beta$, $\lambda$, and $\theta$, then compute $w_1$, $b_1$, $w_2$, and $b_2$ using (35) and (38).
\State Classify testing samples into class $+1$ or $-1$ using (18) or (39).
\end{algorithmic}
\end{algorithm}
We initiate the computational analysis by treating the training dataset $T$ as the initial granular ball set ($GB$). This set $GB$ undergoes a binary split using the $2$-means clustering algorithm, initially resulting in computational complexity of $\mathcal{O}(2m)$. In subsequent iterations, if both resultant granular balls remain impure, they are further divided into four granular balls, maintaining a maximum computational complexity of $\mathcal{O}(2m)$ per iteration. This iterative process continues for a total of $\omega$ iterations.  Consequently, the overall computational complexity of generating granular balls is approximately $\mathcal{O}(2m\omega)$ or less, depending on the purity of the generated granular balls and the number of iterations required.
Suppose that $m_1$ is the number of $+1$ labeled data samples and $m_2$ is the number of $-1$ labeled data samples with $m = m_1 + m_2$. The LSTSVM model requires the calculation of two matrix inverses of order $(m+1)$. However, using the Sherman-Morrison-Woodbury (SMW) formula \cite{golub2013matrix}, the calculation involves solving three inverses of reduced dimensions. Therefore, in the LSTSVM model, the time complexity includes two inversions of size $\mathcal{O}({m_1}^3)$ and one inversion of size $\mathcal{O}({m_2}^3)$ if $m_1 < m_2$. Conversely, if $m_1 \geq m_2$, the complexity involves two inversions of size $\mathcal{O}({m_2}^3)$ and one inversion of size $\mathcal{O}({m_1}^3)$. GBLSTSVM computes the inverses of two matrices with order $(k+1)$, where $k$ represents the total number of granular balls generated on a training dataset $T$. Hence, the total time complexity of the GBLSTSVM model is approximately less than or equal to $\mathcal{O}(2m\omega) + \mathcal{O}({k}^3)$. Given that $\omega$ represents the number of iterations, it follows that $\omega$ is considerably smaller than $m$, and also, $k$ is significantly less than $m$. Thus, $\mathcal{O}(2m\omega) + \mathcal{O}({k}^3) \ll \mathcal{O}({m_1}^3) + \mathcal{O}({m_2}^3)$. Hence, the computational complexity of GBLSTSVM is substantially lower than that of LSTSVM.
The computational complexity of the SMO algorithm is $\mathcal{O}(k)$ to $\mathcal{O}(k^{2.2})$. Therefore, the complexity of each optimization problem in the  LS-GBLSTSVM model falls approximately between $\mathcal{O}(2m\omega) + \mathcal{O}(k)$ and $\mathcal{O}(2m\omega) + \mathcal{O}(k^{2.2}) \ll  \mathcal{O}({m_1}^3) + \mathcal{O}({m_2}^3)$. Therefore, the computational complexity of LS-GBLSTSVM is considerably lower than that of LSTSVM.

\section{Experimental Results and Discussions}
In this section, we assess the efficacy of the proposed GBLSTSVM and LS-GBLSTSVM models. We evaluate their performance against LSTSVM \cite{kumar2009least} and various other baseline models over UCI \cite{dua2017uci} and KEEL \cite{derrac2015keel} benchmark datasets with and without label noise to ensure comprehensive testing. Furthermore, we conduct experiments on NDC datasets \cite{musicant1998ndc}. Moreover, we provide a sensitivity analysis of the hyperparameters and granular ball computing parameters. 
\subsection{Experimental Setup}
To evaluate the performance of the GBLSTSVM and LS-GBLSTSVM models, a series of experiments are conducted. These experiments are carried out on a PC with an Intel(R) Xeon(R) Gold 6226R processor running at 2.90GHz and 128 GB of RAM. The PC is operating on Windows 11 and utilizes Python 3.11. To solve the dual of QPP in GBSVM, the ``QP solvers" function from the CVXOPT package is employed. The dataset is randomly split, with 70\% samples are used for training and 30\% are for testing purposes. The hyperparameters are tuned using the grid search method and five-fold cross-validation.The hyperparameters $c_i$ $(i = 1, 2, 3, 4)$ were tuned within the range $ \lbrace 10^{-5}, 10^{-4}, \cdots, 10^{5} \rbrace $. For the nonlinear case, a Gaussian kernel is utilized, defined as $ K(x_i,x_j) = \exp(\frac{-1}{2\sigma^2}||x_i-x_j||^2)$, where $\sigma$ varied within the range $ \lbrace 2^{-5}, 2^{-4}, \cdots, 2^5 \rbrace$. In the proposed LS-GBLSTSVM model, the values of $c_1$ and $c_2$ are set to be equal, as well as the values of $c_3$ and $c_4$, for both linear and nonlinear cases. 
\subsection{Experments on Real World UCI and KEEL Datasets on the Linear Kernel}
In this subsection, we conduct extensive statistical analyses to compare the proposed GBLSTSVM and LS-GBLSTSVM models with LSTSVM \cite{kumar2009least} along with several other baseline models, namely SVM \cite{cortes1995support}, TSVM \cite{4135685}, and GBSVM \cite{xia2022gbsvm}. To solve the optimization problem associated with GBSVM, we employ the PSO algorithm. Our experimental investigation encompasses diverse scenarios, encompassing both linear and nonlinear cases, and involves meticulous numerical experimentation.
\begin{table*}[ht!]
\caption{Average Accuracy and Average Rank of the baseline models and the proposed models over UCI and KEEL datasets with Linear kernel.}
\label{Average Accuracy and Average Rank of the baseline models and the proposed models over UCI and KEEL datasets with Linear kernel.}
\resizebox{\textwidth}{!} {
\begin{tabular}{cccccccc} \hline
\multicolumn{1}{l}{} & Noise & SVM \cite{cortes1995support} & TSVM \cite{4135685} & GBSVM \cite{xia2022gbsvm} & LSTSVM \cite{kumar2009least}& GBLSTSVM & LS-GBLSTSVM \\ \hline
\multirow{5}{*}{Average ACC} & 0\% & 81.58 & 71.52 & 73.71 & 86.67 & \textbf{88.26} & 86.79 \\
 & 5\% & 80.65 & 81.26 & 77.25 & 85.33 & \textbf{87.50} & 85.59 \\
 & 10\% & 81.70 & 82.83 & 76.63 & 84.82 & \textbf{87.47 }& 85.50 \\
 & 15\% & 80.03 & 80.17 & 73.89 & 84.12 & \textbf{86.04} & 84.93 \\
 & 20\% & 80.21 & 80.55 & 75.76 & 83.79 & \textbf{86.09} & 83.85 \\ \hline
\multirow{5}{*}{Average Rank} & 0\% & 3.88 & 5.26 & 5.15 & 2.59 & \textbf{1.62} & 2.50 \\ 
 & 5\% & 4.46 & 4.37 & 4.97 & 2.94 & \textbf{1.71} & 2.56 \\
 & 10\% & 4.40 & 3.65 & 4.99 & 3.10 & \textbf{1.93} & 2.94 \\
 & 15\% & 4.24 & 4.22 & 5.29 & 2.74 & \textbf{1.68} & 2.84 \\
 & 20\% & 4.29 & 4.19 & 5.16 & 2.72 & \textbf{1.71} & 2.93 \\ \hline
\end{tabular}
}
\end{table*}
We conduct experiments on 34 UCI \cite{dua2017uci} and KEEL \cite{derrac2015keel} benchmark datasets. Table S.2 of Supplementary Material shows the detailed experimental results of every model over each dataset. All the experimental results discussed in this subsection are obtained at a 0\% noise level for both linear and Gaussian kernels. The average accuracy (ACC) and average rank of the linear case are presented in Table \ref{Average Accuracy and Average Rank of the baseline models and the proposed models over UCI and KEEL datasets with Linear kernel.}. The average ACC of the GBLSTSVM model is 88.26\%, while the LS-GBLSTSVM model achieves an average ACC of 86.79\%. On the other hand, the average ACC of the SVM, TSVM, GBSVM, and LSTSVM models are 81.58\%, 71.52\%, 73.71\%, and 86.67\%, respectively. In terms of average ACC, our proposed GBLSTSVM and LS-GBLSTSVM models outperform the baseline SVM, TSVM, GBSVM, and LSTSVM models.  To further evaluate the performance of our proposed models, we employ the ranking method. In this method, each model is assigned a rank for each dataset, with the best-performing model receiving the lowest rank and the worst-performing model receiving the highest rank. The average rank of a model is calculated as the average of its ranks across all datasets. If we consider a set of $M$ datasets, where $l$ models are evaluated on each dataset, we can represent the position of the $s^{th}$ model on the $t^{th}$ dataset as ${r^{t}_s}$. In this case, the average rank of the $s^{th}$ model is calculated as $\mathscr{R}_s = \frac{1}{M}\sum_{t=1}^{M}r^{t}_s$. The average rank of SVM, TSVM, GBSVM, and LSTSVM models are 3.88, 5.26, 5.15, and 2.59, respectively. On the other hand, the average rank of the proposed GBLSTSVM and LS-GBLSTSVM models are 1.62 and 2.50, respectively. Based on the average rank, our proposed models demonstrate a superior performance compared to the baseline models. This indicates that our proposed models exhibit better generalization ability.

To assess the statistical significance of the proposed models, we employ the Friedman test \cite{demvsar2006statistical}. The purpose of this test is to assess the presence of significant disparities among the compared models by examining the average ranks assigned to each model. By evaluating the rankings, we can determine if there are statistically significant differences among the given models. The null hypothesis in this test assumes that all models have the same average rank, indicating an equivalent level of performance. The Friedman test follows the chi-squared distribution ($\chi^2_F$) with $(l-1)$ degrees of freedom and is given by  $\chi^2_F = \frac{12 M}{l(l+1)}\begin{bmatrix} \sum_{s} {\mathscr{R}_s}^2 - \frac{l(l+1)^2}{4}\end{bmatrix}$. The Friedman statistic $F_F$  is given by $F_F = \frac{(M-1)\chi^2_F}{M(l-1)-\chi^2_F}$, where, $F$-distribution has $(l-1)$ and $(l-1)\times(M-1)$ degrees of freedom. For $l = 6$ and $M = 34$, we get  $\chi^2_F = 110.03$ and $F_F = 60.54$ at $5\%$ level significance. From the statistical $F$-distribution table, we find that $F_F(5,165) = 2.2689$. Since $60.54 > 2.2689$, we reject the null hypothesis, indicating a significant statistical difference among the compared models.

To further establish the statistical significance of our proposed GBLSTSVM and LS-GBLSTSVM models with the baseline models, we conduct the Wilcoxon signed rank test \cite{demvsar2006statistical}. This test calculates the differences in accuracy between pairs of models on each dataset. These differences are then ranked in ascending order based on their absolute values, with tied ranks being averaged. Subsequently, the sum of positive ranks ($\mathscr{R}+$) and the sum of negative ranks ($\mathscr{R}-$) are computed. The null hypothesis in this test typically assumes that there is no significant difference between the performances of the models, meaning that the median difference in accuracy is zero. However, if the difference between $\mathscr{R}+$ and $\mathscr{R}-$ is sufficiently large, indicating a consistent preference for one model over the other across the datasets. If the resulting $p$-value from the test is less than $0.05$, then the null hypothesis is rejected. The rejection of the null hypothesis signifies that there exists a statistically significant difference in performance between the compared models.
\begin{table}[ht!]
\centering
\caption{Wilcoxon-signed rank test of the baseline models w.r.t the proposed GBLSTSVM over UCI and KEEL data with the Linear kernel.}
\label{tab:my-table3}
\resizebox{10cm}{!} {
\begin{tabular}{llllc}\hline
Model & $\mathscr{R}+$ & $\mathscr{R}-$ & $p$-value & Null Hypothesis  \\ \hline
SVM \cite{cortes1995support}& 465 & 0 & 0.000001819 & Rejected \\
TSVM \cite{4135685} & 496 & 0 & 0.00000123 & Rejected \\
GBSVM \cite{xia2022gbsvm}& 561 & 0 & 0.0000005639 & Rejected \\
LSTSVM \cite{kumar2009least} & 294.5 & 30.5 & 0.0004013 & Rejected  \\ \hline
\end{tabular}
}
\end{table}
\begin{table}[ht!]
\centering
\caption{Wilcoxon-signed rank test of the baseline models w.r.t. the proposed LS-GBLSTSVM over UCI and KEEL datasets with Linear kernel.}
\label{tab:my-table4}
\resizebox{10cm}{!} {
\begin{tabular}{llllc} \hline
Model & $\mathscr{R}+$ & $\mathscr{R}-$ & $p$-value & Null Hypothesis \\ \hline
SVM \cite{cortes1995support} & 452 & 13 & 0.000006632 & Rejected \\
TSVM \cite{4135685} & 489 & 7 & 0.000002435 & Rejected \\
GBSVM \cite{xia2022gbsvm} & 527 & 1 & 0.0000009169 & Rejected \\
LSTSVM \cite{kumar2009least} & 221.5 & 156.5 & 0.4419 & Not Rejected \\ \hline
\end{tabular}
}
\end{table}

Table \ref{tab:my-table3} presents the results, demonstrating that our proposed GBLSTSVM model outperforms the baseline SVM, TSVM, GBSVM, and LSTSVM models. Furthermore, Table \ref{tab:my-table4} illustrates that the proposed LS-GBLSTSVM model exhibits superior performance compared to the SVM, TSVM, and GBSVM models. The Wilcoxon test strongly suggests that the proposed GBLSTSVM and LS-GBLSTSM models possess a comprehensive statistical advantage over the baseline models.\\

\begin{table}[ht!]
\caption{Pairwise win-tie-loss test of proposed GBLSTSVM and baseline models on UCI and KEEL datasets with linear kernel}
\label{tab:my-table5}
\resizebox{\textwidth}{!} {
\begin{tabular}{lccccc}
\hline
 & SVM \cite{cortes1995support} & TSVM \cite{4135685} & GBSVM \cite{xia2022gbsvm} & LSTSVM \cite{kumar2009least} & GBLSTSVM \\ \hline
TSVM \cite{4135685} & {[}3, 1, 30{]} &  &  &  &  \\ 
GBSVM \cite{xia2022gbsvm}& {[}5, 3, 26{]} & {[}19, 0, 15{]} &  &  &  \\
LSTSVM \cite{kumar2009least} & {[}25, 4, 5{]} & {[}29, 5, 0{]} & {[}31, 1, 2{]} &  &  \\
GBLSTSVM & {[}30, 4, 0{]} & {[}31, 3, 0{]} & {[}33, 1, 0{]} & {[}21, 9, 4{]} &  \\
LS-GBLSTSVM & {[}27, 4, 3{]} & {[}30, 3, 1{]} & {[}31, 1, 2{]} & {[}15, 7, 12{]} & {[}3, 11, 20{]} \\ \hline
\end{tabular}
}
\end{table}
Moreover, we employ a pairwise win-tie-loss sign test. This test is conducted under the assumption that both models are equal and each model wins on $\frac{M}{2}$ datasets, where $M$ denotes the total number of datasets. To establish statistical significance, the model must win on approximately $\frac{M}{2} + 1.96\frac{\sqrt{M}}{2}$ datasets over the other model. In cases where there is an even number of ties between the compared models, these ties are evenly distributed between the models. However, if the number of ties is odd, one tie is disregarded, and the remaining ties are divided among the specified models. In our case, with $M = 34$, if one of the models achieves a minimum of $22.71$ wins, it indicates a significant distinction between the models. 
The results presented in Table \ref{tab:my-table5} clearly show that our proposed models have outperformed the baseline models in the majority of the UCI and KEEL datasets.

\subsection{Experiments on Real World UCI and KEEL Datasets on the Gaussian Kernel}
\begin{table*}[ht!]
\caption{Average Accuracy and Average Rank of the baseline models and the proposed models over UCI and KEEL datasets with Gaussian kernel.}
\label{tab:my-table2}
\resizebox{\textwidth}{!} {
\begin{tabular}{cccccccc} \hline
\multicolumn{1}{l}{} & Noise & SVM \cite{cortes1995support} & TSVM \cite{4135685} & GBSVM \cite{xia2022gbsvm} & LSTSVM \cite{kumar2009least} & GBLSTSVM & LS-GBLSTSVM \\ \hline
\multirow{5}{*}{Average ACC} & 0\% & 75.47 & 82.40 & 78.20 & 76.89 & 83.64 & \textbf{84.55} \\
 & 5\% & 76.02 & 81.07 & 78.42 & 75.03 & \textbf{82.67} & 82.62 \\
 & 10\% & 75.49 & 80.00 & 79.21 & 74.97 &\textbf{83.06} & 82.68 \\
 & 15\% & 77.52 & 82.05 & 80.00 & 75.79 &\textbf{84.05} & 83.15 \\
 & 20\% & 76.63 & 81.93 & 77.86 & 77.17 & \textbf{83.11} & 82.65 \\ \hline
\multirow{5}{*}{Average Rank} & 0\% & 4.32 & 3.24 & 3.99 & 3.72 & 2.90 & \textbf{2.84} \\
 & 5\% & 4.28 & 3.44 & 4.04 & 3.68 & 2.81 & \textbf{2.75} \\
 & 10\% & 4.16 & 3.66 & 3.93 & 3.68 & \textbf{2.69} & 2.88 \\
 & 15\% & 3.91 & 3.44 & 4.28 & 3.78 & \textbf{2.49} & 3.10 \\
 & 20\% & 4.04 & 3.04 & 4.49 & 3.60 & \textbf{2.85} & 2.97 \\ \hline
\end{tabular}
}
\end{table*}
 Supplementary Material's Table 3 demonstrates that our proposed GBLSTSVM and LS-GBLSTSVM models outperform the baseline models in Gaussian kernel space in most of the datasets. Table \ref{tab:my-table2} presents the average accuracy (ACC) and average rank of the proposed GBLSTSVM and LS-GBLSTSVM models, as well as the baseline models using the Gaussian kernel. Our proposed GBSLSTVM and LS-GBLSTSVM models achieve an average accuracy of 83.64\% and 84.55\%, respectively, which is superior to the baseline models. Additionally, the average rank of our proposed GBLSTSVM and LS-GBLSTSVM models is 2.90 and 2.84, respectively, indicating a lower rank compared to the baseline models. This suggests that our proposed models exhibit better generalization ability than the baseline models.

Furthermore, we conduct the Friedman test \cite{demvsar2006statistical} and Wilcoxon signed rank test \cite{demvsar2006statistical} for the Gaussian kernel. For $l = 6$ and $M = 34$, we obtained $\chi^2_F = 16.6279$ and $F_F = 3.5777$ at a significance level of 5\%. Referring to the statistical $F$-distribution table, we find that $F_F(5,165) = 2.2689$. Since $3.5777 > 2.2689$, we reject the null hypothesis. Consequently, there exists a significant statistical difference among the compared models.
\begin{table}[ht!]
\centering
\caption{Wilcoxon-signed rank test of the baseline models w.r.t. the proposed GBLSTSVM over UCI and KEEL datasets with Gaussian kernel.}
\label{tab:my-table6}
\resizebox{10cm}{!} {
\begin{tabular}{lllll} \hline
Model & $\mathscr{R}+$ & $\mathscr{R}-$ & $p$-value & Null Hypothesis \\ \hline
SVM \cite{cortes1995support} & 421 & 75 & 0.0007234 & Rejected \\
TSVM \cite{4135685} & 290 & 238 & 0.6335 & Not Rejected \\
GBSVM \cite{xia2022gbsvm} & 377 & 184 & 0.08628 & Not Rejected \\
LSTSVM \cite{kumar2009least} & 226 & 50 & 0.007777 & Rejected \\ \hline
\end{tabular}
}
\end{table}
\begin{table}[ht!]
\centering
\caption{Wilcoxon-signed rank test of the baseline models w.r.t. the proposed LS-GBLSTSVM over UCI and KEEL datasets with Gaussian kernel.}
\label{tab:my-table7}
\resizebox{10cm}{!} {
\begin{tabular}{lllll} \hline
Model & $\mathscr{R}+$ & $\mathscr{R}-$ & $p$-value & Null Hypothesis \\  \hline
SVM \cite{cortes1995support} & 435 & 93 & 0.00143 & Rejected \\
TSVM \cite{4135685} & 332 & 164 & 0.1018 & Not Rejected \\
GBSVM \cite{xia2022gbsvm} & 409 & 152 & 0.02218 & Rejected \\
LSTSVM \cite{kumar2009least} & 334 & 101 & 0.01213 & Rejected \\ \hline
\end{tabular}
}
\end{table}
\begin{table}[ht!]
\caption{Pairwise win-tie-loss test of proposed and baseline models on UCI and KEEL datasets with Gaussian kernel}
\label{tab:my-table8}
\resizebox{\textwidth}{!} {
\begin{tabular}{lccccc} \hline
\multicolumn{1}{l}{} & SVM \cite{cortes1995support} & TSVM \cite{4135685} & GBSVM \cite{xia2022gbsvm} & \multicolumn{1}{c}{LSTSVM \cite{kumar2009least}} & \multicolumn{1}{c}{GBLSTSVM} \\ \hline
TSVM \cite{4135685} & {[}22, 4, 8{]} & \multicolumn{1}{l}{} & \multicolumn{1}{l}{} &  &  \\
GBSVM \cite{xia2022gbsvm} & {[}17, 1, 16{]} & {[}9, 4, 21{]} & \multicolumn{1}{l}{} &  &  \\
LSTSVM \cite{kumar2009least} & {[}17, 11, 6{]} & {[}15, 1, 18{]} & {[}18, 0, 16{]} &  &  \\
GBLSTSVM & {[}22, 5, 7{]} & {[}18, 2, 14{]} & {[}22, 1, 11{]} & \multicolumn{1}{c}{{[}14, 15, 5{]}} &  \\
LS-GBLSTSVM & {[}22, 5, 7{]} & {[}19, 3, 12{]} & {[}21, 1, 12{]} & \multicolumn{1}{c}{{[}17, 8, 9{]}} & \multicolumn{1}{c}{{[}14, 8, 12{]}} \\ \hline
\end{tabular}
}
\end{table}

Moreover, the Wilcoxon signed test presented in Table \ref{tab:my-table6} for GBLSTSVM and Table \ref{tab:my-table7} for LS-GBLSTSVM demonstrate that our proposed models possess a significant statistical advantage over the baseline models. The pairwise win-tie-loss results presented in Table \ref{tab:my-table8} further emphasize the superiority of our proposed models over the baseline models.
\subsection{Experiments on Real World UCI and KEEL Datasets with Added Label Noise on Linear and Gaussian Kernel}
The proposed GBTSVM and LS-GBTSVM models are experimentally evaluated using UCI and KEEL benchmark datasets. To assess their performance, label noise is introduced at varying levels of 5\%, 10\%, 15\%, and 20\%. The results, presented in Table I and Table II of Supplementary Material, demonstrate the effectiveness of these models compared to baseline models in both linear and nonlinear cases. Throughout the evaluations, the GBLSTSVM and LS-GBLSTSVM models consistently outperformed the baseline models. The proposed GBLSTSVM demonstrates a superior average ACC compared to the baseline models, with an improvement of up to 3\% when increasing the label noise from 5\% to 20\%, for the linear kernel. Similarly, our proposed LS-GBLSTSVM has a better average ACC than the baseline models with linear kernel. The average ACC  of LS-GBLSTSVM at noise levels of 5\%, 10\%, 15\%, and  20\% are 85.59, 85.50, 84.93, and 83.85, respectively, which is higher than all baseline models. Additionally, the proposed models outperform the baseline models in terms of average rank, even when considering various levels of label noise. For the Gaussian kernel, our proposed GBLSTSVM has up to 3\% better average ACC, and  LS-GBLSTSVM has up to 2\% better average ACC compared to baseline models on increasing the levels of label noise from 5\% to 20\%. Also, our proposed models have a lower average rank than the baseline models in increasingly noisy conditions as well.  This can be attributed to the incorporation of granular balls within these models, which exhibit a coarser granularity and possess the ability to mitigate the effects of label noise. The key feature of these granular balls is their strong influence of the majority label within them, effectively reducing the impact of noise points from minority labels on the classification results. This approach significantly enhances the models' resistance to label noise contamination. The consistent superiority of the GBLSTSVM and LS-GBLSTSVM models over the baseline models highlights their potential effectiveness in real-world scenarios where noise is commonly encountered in datasets.

\subsection{Experiment on NDC Datasets}

\begin{table*}[ht!]
\centering
\caption{Accuracy and time of the baseline models and the proposed GBLSTSVM and LS-GBLSTSVM with baseline models on NDC datasets with Linear kernel.}
\label{tab:my-table9}
\resizebox{\textwidth}{!} {
\begin{tabular}{lcccccc} \hline
\multirow{3}{*}{NDC datasets} & SVM \cite{cortes1995support} & TSVM \cite{4135685} & GBSVM \cite{xia2022gbsvm} & LSTSVM \cite{kumar2009least} & GBLSTSVM & LS-GBLSTSVM \\
 & ACC(\%) & ACC(\%) & ACC(\%) & ACC(\%) & ACC(\%) & ACC(\%) \\
 & Time (s) & Time (s) & Time (s) & Time (s) & Time (s) & Time (s) \\ \hline
\multirow{2}{*}{NDC 10K} & 81.64 & 80.78 & 64.97 & 83.34 & \textbf{85.84} & 83.29 \\
 & 310.66 & 209.79 & 1,510.66 & 12.02 & 10.66 & 15.55 \\
\multirow{2}{*}{NDC 50K} & 80.35 & 79.44 & 60.57 & 82.80 & 84.84 & \textbf{84.85} \\
 & 941.11 & 816.81 & 2,809.49 & 54.10 & 30.11 & 39.09 \\
\multirow{2}{*}{NDC 100K} & \multirow{2}{*}{$^c$} & \multirow{2}{*}{ $^c$ } & \multirow{2}{*}{$^d$} & 82.91 & \textbf{85.12} & 84.93 \\
 &  &  &  & 70.16 & 53.47 & 109.86 \\
\multirow{2}{*}{NDC 300K} & \multirow{2}{*}{$^c$} & \multirow{2}{*}{ $^c$ } & \multirow{2}{*}{$^d$}& 83.21 & \textbf{84.86} & 83.93 \\
 &  &  &  & 124.44 & 112.41 & 159.27 \\
\multirow{2}{*}{NDC 500K} & \multirow{2}{*}{$^c$}& \multirow{2}{*}{ $^c$ } & \multirow{2}{*}{$^d$} & 83.14 & \textbf{85.90} & 83.79 \\
 &  &  &  & 199.72 & 165.98 & 184.03 \\
\multirow{2}{*}{NDC 1m} & \multirow{2}{*}{$^c$} & \multirow{2}{*}{ $^c$ } & \multirow{2}{*}{$^d$} & 83.07 & \textbf{84.75} & 83.94 \\
 &  &  &  & 301.42 & 221.76 & 265.91 \\
\multirow{2}{*}{NDC 3m} & \multirow{2}{*}{$^c$} & \multirow{2}{*}{ $^c$ }& \multirow{2}{*}{$^d$} & 83.02 & \textbf{84.56} & 83.61 \\
 &  &  &  & 357.24 & 267.65 & 291.51 \\
\multirow{2}{*}{NDC 5m} & \multirow{2}{*}{$^c$} & \multirow{2}{*}{ $^c$ }& \multirow{2}{*}{$^d$} & 83.10 & 84.30 & \textbf{84.99} \\
 &  &  &  & 406.63 & 316.89 & 499.65 \\ \hline
\multicolumn{7}{l}{\begin{tabular}[c]{@{}l@{}} $^c$ Terminated because of out of memory. \\  $^d$ Experiment is terminated because of the out of bound issue shown by the PSO algorithm.
\end{tabular}}
\end{tabular} } 
\end{table*}
The previous comprehensive analyses have consistently shown the superior performance of the proposed GBLSTSVM and LS-GBLSTSVM models compared to the baseline models across the majority of UCI and KEEL benchmark datasets. Furthermore, we conduct an experiment using the NDC datasets \cite{musicant1998ndc} to highlight the enhanced training speed and scalability of our proposed models. For this, all hyperparameters are set to $10^{-5}$, which is the lowest value among the specified ranges. These NDC datasets' sample sizes vary from 10k to 5m with 32 features. The results presented in Table \ref{tab:my-table9} show the efficiency and scalability of the proposed GBLSTSVM and LS-GBLSTSVM models. Across the NDC datasets, our models consistently outperform the baseline models in terms of both accuracy and training times, thus confirming their robustness and efficiency, particularly when dealing with large-scale datasets. In the context of ACC, our GBLSTSVM model demonstrates superior accuracy, with an increase of up to 3\% when the NDC dataset scale is expanded from 10k to 5m. Additionally, GBLSTSVM demonstrates reduced training time across all ranges of NDC datasets compared to LSTSVM. Specifically, for the NDC 5m dataset, the LS-GBLSTSVM model achieves an impressive accuracy of 84.99\%, which stands as the highest accuracy. The experimental results demonstrate a significant reduction of 100 to 1000 times in the training duration of the GBLSTSVM and LS-GBLSTSVM models compared to the baseline models.  This exceptional decrease in training time can be attributed to the significantly lower count of generated granular balls on a dataset in comparison to the total number of samples.

\subsection{Performance Comparison: Dimensionality Reduction Perspective}
Granular computing generates granular balls, which serve as compact and high-level representations of the data, capturing the intrinsic patterns and distribution of the dataset. Instead of relying on individual training data points, this approach abstracts the dataset into granular balls that act as a summary of data clusters. These granular balls encapsulate important statistical properties, such as centers and radii, representing the spread and variability of data within each cluster. By replacing the original training data points with granular balls, the computational burden of processing large datasets is significantly reduced. This transformation leads to a substantial reduction in the number of data points involved in the training process, effectively simplifying the problem space. This reduction can be interpreted as a form of dimensionality reduction, not in the traditional sense of reducing the number of features, but by reducing the effective size of the dataset used for computation. Despite the reduction in data points, the granular balls retain critical information about the data's structure and relationships, ensuring that the model has access to the most relevant patterns for effective learning.
\begin{table}[ht!]
\centering
\caption{Average accuracy comparison of SLCE, linear GBSVM, the proposed linear GBLSTSVM, and linear LS-GBLSTSVM on UCI and KEEL Datasets.}
\label{dimensionality}
\resizebox{\textwidth}{!} {
\begin{tabular}{cccccc}
\hline
\multirow{7}{*}{Average ACC} & Noise & GBSVM \cite{xia2022gbsvm} & SLCE \cite{ghosh2024linear} & GBLSTSVM & LS-GBLSTSVM \\
\hline
 & 0\% & 73.71 & 80.26 & \textbf{88.26} & 86.79 \\
 & 5\% & 77.25 & 76.18 & \textbf{87.50} & 85.59 \\
 & 10\% & 76.63 & 72.52 & \textbf{87.47} & 85.50 \\
 & 15\% & 73.89 & 69.91 & \textbf{86.04} & 84.93 \\
 & 20\% & 75.76 & 65.43 & \textbf{86.09} & 83.85 \\
 \hline 
\end{tabular}}
\end{table}
To provide a more comprehensive comparison of the proposed models, we compare it with the linear centroid encoder for supervised principal component analysis (SLCE) \cite{ghosh2024linear}. SLCE is a linear dimensionality reduction technique that leverages class centroids, rather than labels, to incorporate supervision into the learning process. This approach enables SLCE to capture class-specific structures effectively, making it a relevant benchmark for evaluating the performance of the proposed model in terms of dimensionality reduction and supervised learning. The performance of SLCE on UCI and KEEL datasets at various levels of noise along with the standard deviation (std dev) is presented in Table 4 of the Supplementary file. The experiments for SLCE were conducted under the same experimental setup and conditions as described in \cite{ghosh2024linear} and the training and testing datasets were divided in the ratio of $70:30$.

Table \ref{dimensionality} presents the average ACC of the compared models on UCI and KEEL datasets across various levels of noise. The results clearly demonstrate that the proposed models outperform the baseline models, GBSVM and SLCE. Although GBSVM incorporates a similar dimensionality reduction strategy through granular computing, the proposed models exhibit significantly enhanced performance. This improvement highlights their superior capability to capture and leverage the structural information within the data, leading to more effective and robust classification outcomes. 

 For a comprehensive evaluation of the models, experiments were conducted on publicly available biological datasets\footnote{\url{https://jundongl.github.io/scikit-feature/datasets.html}} comprising two classes, as the proposed models are specifically designed for binary classification tasks. Table \ref{tab} provides a detailed description of the biological datasets used in the experiments, while Table \ref{tab:dataset_description} presents the performance results of the proposed Linear GBLSTSVM and Linear LS-GBLSTSVM models compared to the baseline models, GBSVM and SLCE. The results in Table \ref{tab:dataset_description} clearly demonstrate that the proposed models consistently outperform the baseline models, showcasing their superior effectiveness and robustness. Additional experimental details, including the hyperparameters and standard deviations, are provided in Table 1 of the supplementary file.

\begin{table}[ht!]
\centering
\caption{Description of Biological Datasets}
\label{tab}
\begin{tabular}{lcc}
\hline
\textbf{Dataset Name} & \textbf{Samples} & \textbf{Features} \\ \hline
ALLAML               & 72               & 7129                            \\ \hline
colon                & 62               & 2000                             \\ \hline
leukemia             & 72               & 7070                             \\ \hline
Prostate\_GE         & 102              & 5966                              \\ \hline
SMK\_CAN\_187        & 187              & 19993                            \\ \hline
\end{tabular}
\end{table}

\begin{table}[h]
\centering
\caption{Performance Results on Biological Datasets}
\label{tab:dataset_description}
\begin{tabular}{ccccc}
\hline
Model         & GBSVM \cite{xia2022gbsvm} & SLCE \cite{ghosh2024linear}   & GBLSTSVM & LS-GBLSTSVM \\
Dataset       & ACC(\%) & ACC(\%) & ACC(\%)  & ACC(\%)     \\
\hline
ALLAML        & 78.12     & \textbf{92.91}  &  89.11    & 90.00          \\
colon         & 70.68      & 80.84  & \textbf{90.77}       & 77.17         \\
leukemia      & 76.19     & \textbf{96.18}  & 81.50         & 77.56           \\
Prostate\_GE  & 80.28     & 87.61  & \textbf{92.72}         & 74.60        \\
SMK\_CAN\_187 & 62.54      & 69.26  &   \textbf{73.02 }     & 68.25          \\
\hline
\end{tabular} 
\end{table}

\subsection{Sensitivity Analysis of Hyperparameters}

\begin{figure*}[ht!]
\begin{minipage}{.24\linewidth}
\centering
\subfloat[yeast-0-2-5-6 vs 3-7-8-9]{\label{fig:1a}\includegraphics[scale=0.35]{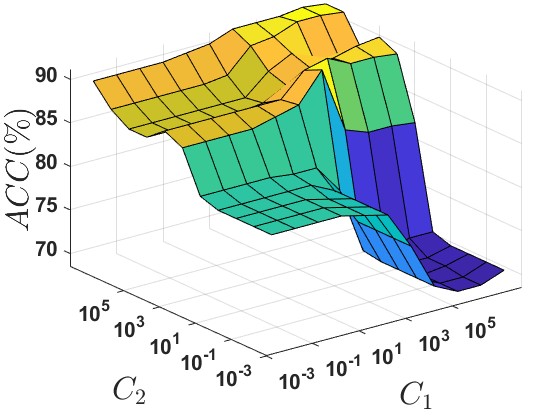}}
\end{minipage}
\begin{minipage}{.5\linewidth}
\centering
\subfloat[Haber]{\label{fig:1b1}\includegraphics[scale=0.35]{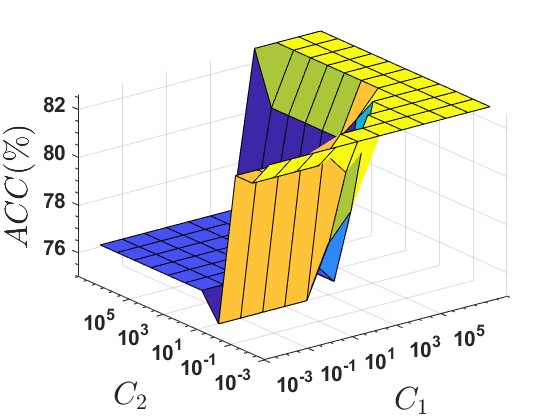}}
\end{minipage}
\begin{minipage}{.5\linewidth}
\centering
\subfloat[Monks3]{\label{fig:1d1}\includegraphics[scale=0.35]{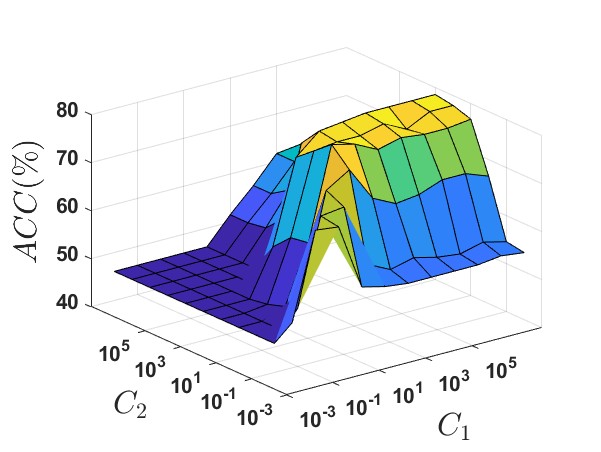}}
\end{minipage}
\begin{minipage}{.5\linewidth}
\centering
\subfloat[Spambase]{\label{fig:1e1}\includegraphics[scale=0.35]{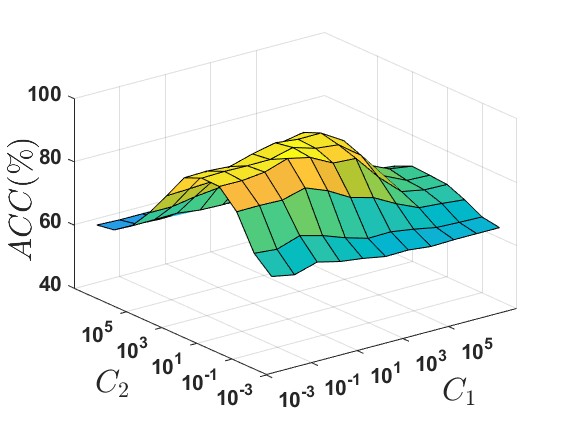}}
\end{minipage}
\caption{The effect of hyperparameter $(c_1, c_2)$ tuning on the accuracy (ACC) of some UCI and KEEL datasets on the performance of linear GBLSTSVM.}
\label{}
\end{figure*}

\begin{figure*}[ht!]
\begin{minipage}{.5\linewidth}
\centering
\subfloat[breast cancer wisc prog]{\label{fig:1a1}\includegraphics[scale=0.35]{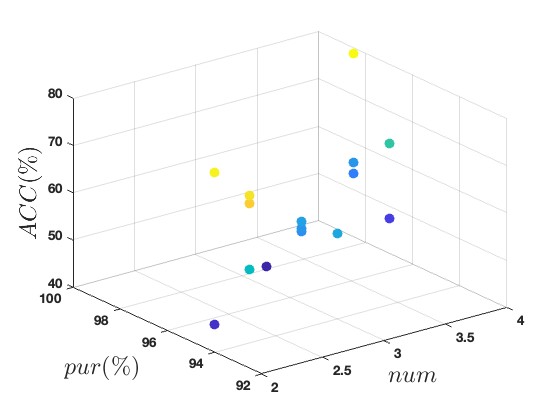}}
\end{minipage}
\begin{minipage}{.5\linewidth}
\centering
\subfloat[haberman]{\label{fig:1b}\includegraphics[scale=0.35]{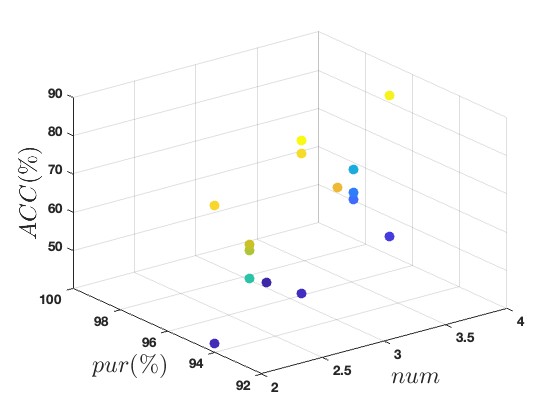}}
\end{minipage}
\begin{minipage}{.5\linewidth}
\centering
\subfloat[Monks2]{\label{fig:1d}\includegraphics[scale=0.35]{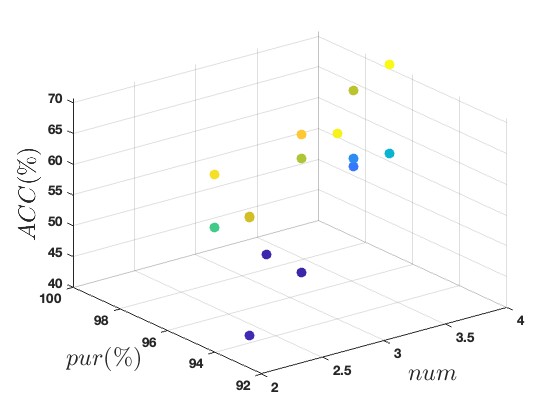}}
\end{minipage}
\begin{minipage}{.5\linewidth}
\centering
\subfloat[Spambase]{\label{fig:1e}\includegraphics[scale=0.35]{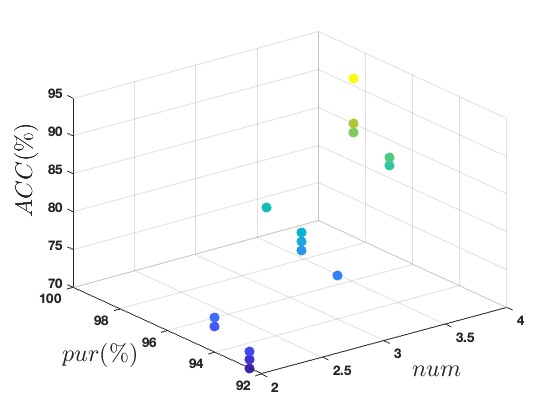}}
\end{minipage}
\caption{The effect of granular parameter $(num, pur)$ tuning on the accuracy (ACC) of some UCI and KEEL datasets on the performance of linear GBLSTSVM.}
\label{}
\end{figure*}

To thoroughly understand the subtle effects of the hyperparameters on the model's generalization ability, we systematically explore the hyperparameter space by varying the values of $c_1$ and $c_2$. This exploration allows us to identify the configuration that maximizes predictive accuracy and enhances the model's resilience to previously unseen data. The graphical representations in Fig. 4 provide visual insights into the impact of parameter tuning on the accuracy (ACC) of our GBLSTSVM model for linear case. These visuals demonstrate an apparent variation in the model's accuracy across a range of $c_1$ and $c_2$ values, highlighting the sensitivity of our model's performance to these hyperparameters. In Fig. 4(a), it is evident that lower values of $c_1$ combined with higher values of $c_2$ result in improved accuracy. Similarly, Fig. 4(d) shows that optimal accuracy is achieved when both $c_1$ and $c_2$ are set to mid-range values. It has been observed that lower values of $c_1$ and higher values of $c_2$ give the best generalization performance.
{\begin{table}[ht!]
\caption{Performance of the proposed Linear GBLSTSVM across varying purities, showcasing the relationship between the number of Granular Balls and the resulting accuracies.}
\label{tab:my-table10}
\resizebox{\textwidth}{!} {
\begin{tabular}{ccccccccc} \hline
$pur$                          & 1       & 0.97    & 0.94    & 0.91    & 0.88    & 0.85    & 0.82    & 0.79    \vspace{2mm} \\ \hline \vspace{-3mm} \\ 
\multirow{2}{*}{Dataset}       & ACC   & ACC   & ACC   & ACC   & ACC   & ACC   & ACC   & ACC   \\
                               & $n(GB)$ & $n(GB)$ & $n(GB)$ & $n(GB)$ & $n(GB)$ & $n(GB)$ & $n(GB)$ & $n(GB)$ \vspace{2mm} \\ \hline \hline \vspace{-3mm} \\ 
\multirow{2}{*}{spambase}      & 89.21   & 90.88   & 86.31   & 83.13   & 79.36   & 81.03   & 78.13   & 78.86   \\
                               & 400     & 316     & 277     & 201     & 180     & 134     & 107     & 99      \\ \hline
\multirow{2}{*}{musk\_1}       & 83.22   & 78.32   & 81.12   & 79.72   & 79.72   & 80.42   & 79.72   & 78.52   \\
                               & 59   & 63   & 60  & 57   & 51   & 48   & 37   & 40  \\ \hline
\multirow{2}{*}{tic\_tac\_toe} & 99.65   & 99.65   & 75.35   & 97.22   & 99.65   & 99.65   & 99.65   & 99.65   \\
                               & 108     & 113     & 105     & 100     & 93      & 73      & 76      & 57      \\ \hline
\multirow{2}{*}{monks\_3}      & 80.24   & 83.23   & 81.44   & 79.04   & 80.84   & 80.84   & 82.04   & 79.04   \\
                               & 69      & 67      & 65      & 67      & 74      & 60      & 56      & 54      \\ \hline
\multirow{2}{*}{vehicle1}      & 83.46   & 84.25   & 84.25   & 83.07   & 76.77   & 82.28   & 84.25   & 79.13   \\
                               & 96      & 94      & 93      & 75      & 69      & 61      & 49      & 31       \\ \hline
\end{tabular} 
} 
\end{table}
{\begin{table}[ht!]
\caption{Performance of the proposed Linear LS-GBLSTSVM across varying purities, showcasing the relationship between the number of Granular Balls and the resulting accuracies.}
\label{tab:my-table11}
\resizebox{14cm}{!} {
\begin{tabular}{ccccccccc} \hline
$pur$                           & 1       & 0.97    & 0.94    & 0.91    & 0.88    & 0.85    & 0.82    & 0.79    \vspace{2mm} \\ \hline \vspace{-3mm} \\
\multirow{2}{*}{Dataset}        & ACC   & ACC   & ACC   & ACC   & ACC   & ACC   & ACC   & ACC   \\
                                & $n(GB)$ & $n(GB)$ & $n(GB)$ & $n(GB)$ & $n(GB)$ & $n(GB)$ & $n(GB)$ & $n(GB)$ \vspace{2mm} \\ \hline \hline \vspace{-3mm} \\
\multirow{2}{*}{musk\_1}        & 79.02   & 75.52   & 79.02   & 74.13   & 77.62   & 72.73   & 78.32   & 76.22   \\
                                & 59      & 63      & 60      & 57      & 51      & 48      & 37      & 40      \\ \hline
\multirow{2}{*}{tic\_tac\_toe}  & 97.92   & 100.00  & 90.63   & 95.14   & 97.22   & 98.26   & 99.65   & 91.32   \\
                                & 108     & 113     & 105     & 100     & 93      & 73      & 76   & 57  \\ \hline
\multirow{2}{*}{monks\_3}       & 67.07   & 68.26   & 60.48   & 55.09   & 76.05   & 70.66   & 64.07   & 61.68   \\
                                & 69      & 67      & 65      & 67      & 74      & 60      & 56      & 54      \\ \hline
\multirow{2}{*}{monks\_2}       & 53.04   & 50.28   & 48.07   & 64.14   & 58.01   & 44.75   & 51.38   & 64.09   \\
                                & 85      & 92      & 88      & 89      & 77      & 72      & 81      & 62      \\ \hline
\multirow{2}{*}{breast\_cancer} & 60.47   & 74.41   & 66.28   & 37.21   & 43.02   & 55.81   & 62.79   & 60.47   \\
                                & 41      & 40      & 39      & 40      & 36      & 35      & 31      & 24      \\ \hline
\end{tabular} 
} 
\end{table}
\subsection{Sensitivity Analysis of Granular Parameters}
In the context of granular computing, as elaborated in Section II (B), we can ascertain the minimum number of granular balls to be generated on the training dataset $T$, denoted as $num$. For our binary classification problem, we establish the minimum value for $num$ at 2. Hence, our goal is to generate at least two granular balls for each dataset. The purity $(pur)$ of a granular ball is a crucial characteristic. By adjusting the purity level of the granular balls, we can simplify the distribution of data points in space and effectively capture the data points distribution using these granular balls. To analyze the impact of $num$ and $pur$ on the generalization performance of GBLSTSVM, we have tuned $num$ within the range of $\lbrace2, 3, 4\rbrace$, and  $pur$ within the range of $\lbrace 92.5, 94.0, 95.5, 97.0,  98.5 \rbrace$. The visual depictions in Fig. 5 offer valuable visualizations of how this tuning affects the accuracy (ACC) of our GBLSTSVM model. Careful examination of these visuals depicts that there exists an optimal value of $num$ and $pur$ using which our proposed GBLSTSVM mode gives the optimal generalization performance. In the case of Fig. 5(a), when both $pur$ and $num$ are increased simultaneously, a significant rise in ACC can be observed. This suggests that the accuracy of our proposed model improves as the purity increases and the minimum number of granular balls generated increases. Similar patterns can be observed in other figures in Fig. 5. This aligns with the principle of granular computing. As the value of $num$ rises, the minimum count of granular balls required to cover our sample space also increases. With an increase in $pur$, these granular balls divide even more, resulting in a greater generation of granular balls. This process effectively captures the data patterns, ultimately leading to the best possible generalization performance. Tables \ref{tab:my-table10} and \ref{tab:my-table11} illustrate the variation in ACC for linear GBLSTSVM and LS-GBLSTSVM model and the number of granular balls generated  ($n(GB)$) while tuning purity levels across various UCI and KEEL datasets when $num$ is fixed at 2.
\section{Conclusion}
\label{7}

In this paper, we have proposed two novel models, the granular ball least square twin support vector machine (GBLSTSVM) and the large-scale granular ball least square twin support vector machine (LS-GBLSTSVM), by incorporating the concept of granular computing in LSTSVM. This incorporation enables our proposed models to achieve the following: (i) Robustness: Both GBLSTSVM and LS-GBLSTSVM demonstrate robustness due to the coarse nature of granular balls, making them less susceptible to noise and outliers.
(ii) Efficiency: The efficiency of GBLSTSVM and LS-GBLSTSVM stems from the significantly lower number of coarse granular balls compared to finer data points, enhancing computational efficiency.
(iii) Scalability: Our proposed models are well suited for large-scale problems, primarily due to the significantly reduced number of generated granular balls compared to the total training data points. The proposed LS-GBLSTSVM model demonstrates exceptional scalability since it does not necessitate matrix inversion for determining optimal parameters. This is evidenced by experiments conducted on the NDC dataset, showcasing their ability to handle large-scale datasets effectively. An extensive series of experiments and statistical analyses have supported the above claims, including ranking schemes, the Friedman Test, the Wilcoxon Signed Rank Test, and the win-tie-loss sign test. The key findings from our experiments include:
(i)  Both linear and nonlinear versions of GBLSTSVM and LS-GBLSTSVM demonstrate superior efficiency and generalization performance compared to baseline models, with an average accuracy improvement of up to 15\%. (ii) When exposed to labeled noise in the UCI and KEEL datasets, our models exhibit exceptional robustness, achieving up to a 10\% increase in average accuracy compared to baseline models under noisy conditions. (iii) Evaluating our proposed models on NDC datasets ranging from 10k to 5m samples underscores their scalability, surpassing various baseline models in training speed by up to 1000 times, particularly beyond the NDC-50k threshold, where memory limitations often hinder baseline models. These findings collectively highlight the effectiveness, robustness, and scalability of our proposed GBLSTSVM and LS-GBLSTSVM models, particularly in handling large-scale and noisy datasets. From the perspective of dimensionality reduction, our proposed model demonstrates superior performance across UCI and KEEL datasets, as well as biological datasets, highlighting its effectiveness in handling diverse data structures while maintaining robust classification accuracy. While our proposed models have demonstrated outstanding performance in binary classification problems, their evaluation on multiclass problems has not been conducted. A crucial area for future research would involve adapting these models to be suitable for multi-class problems.

\section*{Acknowledgment}
This project is supported by the Indian government through grants from the Department of Science and Technology (DST) and the Ministry of Electronics and Information Technology (MeitY). The funding includes DST/NSM/R\&D\_HPC\_Appl/2021/03.29 for the National Supercomputing Mission and MTR/2021/000787 for the Mathematical Research Impact-Centric Support (MATRICS) scheme. Furthermore, Md Sajid's research fellowship is funded by the Council of Scientific and Industrial Research (CSIR), New Delhi, under the grant 09/1022(13847)/2022-EMR-I. The authors express their gratitude for the resources and support provided by IIT Indore.
 
\bibliography{refer.bib}
\bibliographystyle{unsrtnat}

\clearpage
\section*{Supplementary Material}

\renewcommand{\thesection}{S.1}
\section{Classification accuracies of the proposed GBLSTSVM and LS-GBLSTSVM models alongside baseline models with linear and Gaussian kernels under varying levels of noise, and comparative analysis on UCI, KEEL, and biological datasets}

Here, we provide a comprehensive analysis, comparing the performance of the proposed models with baseline models employing linear and Gaussian kernels under varying levels of label noise (0\%, 5\%, 10\%, 15\%, and 20\%). The results are presented in Table 2 and Table 3. Notably, the proposed models consistently outperform the baseline models, achieving the highest average accuracy (ACC) and the lowest average rank across all noise levels. For ease of reference, the best results are highlighted in boldface. 

Additionally, we present the performance of the linear centroid encoder for supervised principal component analysis (SLCE) \cite{ghosh2024linear} on UCI and KEEL datasets at various noise levels in Table 4. Furthermore, results on biological datasets are provided in Table 1, showcasing the robustness and effectiveness of the proposed models in diverse scenarios.

\renewcommand{\thetable}{S.1}
\begin{table}[h]
\centering
\caption{Performance Results on Biological Datasets}
\label{tab:dataset_descriptions}
\begin{tabular}{ccccc}
\hline
Model         & GBSVM \cite{xia2022gbsvm} & SLCE \cite{ghosh2024linear}   & GBLSTSVM & LS-GBLSTSVM \\
       & ACC(\%) & ACC(\%) & ACC(\%)  & ACC(\%)     \\
Dataset& $(c_1)$ &  & $(c_1, c_2)$&   $(c_1, c_2)$ \\
\hline
ALLAML        & 78.12     & \textbf{92.91}  &  89.11    & 90.00          \\
& (0.001) &  & (0.1, 10) & (0.001, 100) \\ 
colon         & 70.68      & 80.84  & \textbf{90.77}       & 77.17         \\
& (10) &  &  (10, 1000) & (0.1, 0.0001) \\
leukemia      & 76.19     & \textbf{96.18}  & 81.50         & 77.56           \\
& (10000)&  &   (1, 100) & (0.001, 10) \\
Prostate\_GE  & 80.28     & 87.61  & \textbf{92.72}         & 74.60        \\
& (0.1) &  &(1000, 1) & (100, 0.1) \\
SMK\_CAN\_187 & 62.54      & 69.26  &   \textbf{73.02 }     & 68.25          \\
& (10000) &  & (10000, 10) & (0.0001, 1000)\\
\hline
\end{tabular} 
\end{table}

\renewcommand{\thetable}{S.2}
\begin{table*}[hb]
\caption{Performance comparison of the proposed GBLSTSVM and LS-GBLSTSVM along with the baseline models based on classification accuracy using linear kernel for UCI and KEEL datasets.}
\label{tab:my-table1000}
\centering
\resizebox{\textwidth}{!}{
\begin{tabular}{cccccccc} \hline
Model & \multicolumn{1}{l}{} & SVM \cite{cortes1995support} & TSVM \cite{4135685} & GBSVM \cite{xia2022gbsvm} & LSTSVM \cite{kumar2009least} & GBLSTSVM & LS-GBLSTSVM \\
\multirow{2}{*}{Dataset} & \multirow{2}{*}{Noise} & ACC ($\%$) & ACC ($\%$) & ACC ($\%$) & ACC ($\%$) & ACC ($\%$) & ACC ($\%$) \\
 &  & ($c_1$) & ($c_1, c_2$) & ($c_1$) & ($c_1, c_2$) & ($c_1, c_2$) & ($c_1, c_2$) \\ \hline
\multicolumn{1}{c}{\multirow{10}{*}{acute\_nephritis}} & 0$\%$ & \textbf{\textbf{100.00}} & \textbf{100.00 }& 90.33 & \textbf{\textbf{100.00}} & \textbf{\textbf{100.00}} & \textbf{\textbf{100.00}} \\
\multicolumn{1}{c}{\multirow{10}{*}{(120 $\times$ 6)}}&
\multicolumn{1}{l}{} & (0.1) & ( 0.01, 0.01) & (100000) & ( 0.001, 0.001) & ( 0.00001, 1000 ) & (0.00001, 0.00001) \\
\multicolumn{1}{c}{} & 5$\%$ & \textbf{\textbf{100.00}} & \textbf{\textbf{100.00}} & 90.33 & \textbf{\textbf{100.00}} & \textbf{\textbf{100.00}} & \textbf{\textbf{100.00}} \\
\multicolumn{1}{c}{} & \multicolumn{1}{l}{} & (0.1) & (0.1, 1) & (100000) & (0.1, 1) & (0.00001, 0.01) & (0.00001, 0.00001) \\
\multicolumn{1}{c}{} & 10$\%$ & \textbf{\textbf{100.00}} & \textbf{\textbf{100.00}} & 94.44 & \textbf{\textbf{100.00}} & \textbf{\textbf{100.00}} & \textbf{\textbf{100.00}} \\
\hline
 \end{tabular}}
\end{table*}

\begin{table*}[!h]
\caption*{\text{Table S.2:} (Continued)}
\label{tab:my-table20}
\centering
\resizebox{0.90\linewidth}{!}{
\begin{tabular}{cccccccc} \hline
Model & \multicolumn{1}{l}{} & SVM \cite{cortes1995support} & TSVM \cite{4135685} & GBSVM \cite{xia2022gbsvm} & LSTSVM \cite{kumar2009least} & GBLSTSVM & LS-GBLSTSVM \\
\multirow{2}{*}{Dataset} & \multicolumn{1}{c}{\multirow{2}{*}{Noise}} & ACC ($\%$) & ACC ($\%$) & ACC ($\%$) & ACC ($\%$) & ACC ($\%$) & ACC ($\%$) \\
 & \multicolumn{1}{c}{} & ($c_1$) & ($c_1, c_2$) & ($c_1$) & ($c_1, c_2$) & ($c_1, c_2$) & ($c_1, c_2$) \\  \hline
 \multirow{10}{*}{crossplane150} & 0$\%$ & 55.56 & \textbf{100.00} & 72.22 & \textbf{100.00} & \textbf{100.00} & 97.78 \\
\multicolumn{1}{c}{\multirow{10}{*}{(150 $\times$ 2)}}
 & \multicolumn{1}{l}{} & (0.01) & (0.00001, 0.00001) & (100000) & (0.00001, 0.00001) & (0.00001, 0.00001) & (0.0001, 0.00001) \\
 & 5$\%$ & 55.56 & 95.56 & 86.67 & 95.56 & \textbf{100.00} & 80.00 \\
 & \multicolumn{1}{l}{} & (0.01) & (0.00001, 0.00001) & (0.00001) & (0.00001, 0.00001) & (0.00001, 0.00001) & (0.00001, 0.1) \\
 & 10$\%$ & 53.33 & 95.56 & 86.67 & 75.56 & \textbf{100.00} & 97.78 \\
\hline
 \end{tabular}
}
\end{table*}

\begin{table*}[!h]
\caption*{\text{Table S.2:} (Continued)}
\label{tab:my-table30}
\centering
\resizebox{0.95\linewidth}{!}{
\begin{tabular}{cccccccc} \hline
Model & \multicolumn{1}{l}{} & SVM \cite{cortes1995support} & TSVM \cite{4135685} & GBSVM \cite{xia2022gbsvm} & LSTSVM \cite{kumar2009least} & GBLSTSVM & LS-GBLSTSVM \\
\multirow{2}{*}{Dataset} & \multicolumn{1}{c}{\multirow{2}{*}{Noise}} & ACC ($\%$) & ACC ($\%$) & ACC ($\%$) & ACC ($\%$) & ACC ($\%$) & ACC ($\%$) \\
 & \multicolumn{1}{c}{} & ($c_1$) & ($c_1, c_2$) & ($c_1$) & ($c_1, c_2$) & ($c_1, c_2$) & ($c_1, c_2$) \\  \hline
\multirow{10}{*}{glass5} & \multicolumn{1}{c}{0$\%$} & 92.92 & 89.00 & 80.46 & 95.38 & \textbf{96.92} & \textbf{96.92} \\ 
\multicolumn{1}{c}{\multirow{10}{*}{(214 $\times$9)}}
 &  & (0.00001) & (0.1, 1) & (100000) & (0.00001, 0.1) & (10000, 1000) & (0.00001, 0.01) \\
 & \multicolumn{1}{c}{5$\%$} & 92.92 & 92.46 & 86.15 & \textbf{96.92} & \textbf{96.92} &\textbf{ 96.92} \\
 &  & (0.00001) & (1, 1) & (10) & (0.1, 0.1) & (0.00001, 0.00001) & (0.00001, 0.01) \\
 & \multicolumn{1}{c}{10$\%$} & 92.92 & 92.46 & \textbf{96.92} & \textbf{96.92} & \textbf{96.92} & \textbf{96.92} \\
\hline
\end{tabular}
}
\end{table*}

\begin{table*}[ht!]
\caption*{\text{Table S.2:} (Continued)}
\label{tab:my-table40}
\centering
\resizebox{0.95\linewidth}{!}{
\begin{tabular}{cccccccc} \hline
Model & \multicolumn{1}{l}{} & SVM \cite{cortes1995support} & TSVM \cite{4135685} & GBSVM \cite{xia2022gbsvm} & LSTSVM \cite{kumar2009least} & GBLSTSVM & LS-GBLSTSVM \\
\multirow{2}{*}{Dataset} & \multicolumn{1}{c}{\multirow{2}{*}{Noise}} & ACC ($\%$) & ACC ($\%$) & ACC ($\%$) & ACC ($\%$) & ACC ($\%$) & ACC ($\%$) \\
 & \multicolumn{1}{c}{} & ($c_1$) & ($c_1, c_2$) & ($c_1$) & ($c_1, c_2$) & ($c_1, c_2$) & ($c_1, c_2$) \\  \hline
\multirow{10}{*}{monks\_2} & \multicolumn{1}{c}{0$\%$} & 62.98 & 62.43 & 62.54 & 62.98& 62.98& \textbf{64.64} \\
 \multicolumn{1}{c}{\multirow{10}{*}{(601 $\times$6)}}
 &  & (0.00001) & (0.0001, 0.00001) & (1000) & (0.0001, 0.00001) & (100, 10000) & (0.00001, 0.00001) \\
 & \multicolumn{1}{c}{5$\%$} & 61.78 & 61.98 & 59.72 & \textbf{62.98} & \textbf{62.98} & \textbf{62.98} \\
 &  & (0.00001) & (0.0001, 0.00001) & (10) & (0.0001, 0.00001) & (10, 100) & (1, 10000) \\
 & \multicolumn{1}{c}{10$\%$} & 60.98 & 62.98 & 60.49 & 62.98 & 62.98 & \textbf{64.09} \\
\hline
\end{tabular}
}
\end{table*}

\begin{table*}[ht!]
\caption*{\text{Table S.2:} (Continued)}
\label{tab:my-table50}
\centering
\resizebox{1\linewidth}{!}{
\begin{tabular}{cccccccc} \hline
Model & \multicolumn{1}{l}{} & SVM \cite{cortes1995support} & TSVM \cite{4135685} & GBSVM \cite{xia2022gbsvm} & LSTSVM \cite{kumar2009least} & GBLSTSVM & LS-GBLSTSVM \\
\multirow{2}{*}{Dataset} & \multicolumn{1}{c}{\multirow{2}{*}{Noise}} & ACC ($\%$) & ACC ($\%$) & ACC ($\%$) & ACC ($\%$) & ACC ($\%$) & ACC ($\%$) \\
 & \multicolumn{1}{c}{} & ($c_1$) & ($c_1, c_2$) & ($c_1$) & ($c_1, c_2$) & ($c_1, c_2$) & ($c_1, c_2$) \\  \hline
\multirow{10}{*}{vehicle1 } & \multicolumn{1}{c}{0$\%$} & 76.38 & 59.44 & 73.62 & 77.95 & \textbf{83.46} & 77.17 \\
\multicolumn{1}{c}{\multirow{10}{*}{\hspace{-0.2cm} (846 $\times$18)}}
 &  & (0.00001) & (10, 1) & (10) & (100, 10) & (1, 1) & (0.01, 0.00001) \\
 & \multicolumn{1}{c}{5$\%$} & 76.38 & 78.74 & 73.62 & 77.95 & \textbf{81.50} & 77.56 \\
 &  & (0.00001) & (10, 1) & (10) & (100, 10) & (0.01, 0.1) & (0.01, 0.00001) \\
 & \multicolumn{1}{c}{10$\%$} & 76.38 & 81.50 & 71.26 & \textbf{83.86} & 83.46 & 77.56 \\
\hline
\end{tabular}
}
\end{table*}

\renewcommand{\thetable}{S.3}
\begin{table*}[ht!]
\caption{Performance comparison of the proposed GBLSTSVM and LS-GBLSTSVM along with the baseline models based on classification accuracy using Gaussian kernel for UCI and KEEL datasets.}
\label{tab:my-table60}
\centering
\resizebox{1.05\linewidth}{!}{
\begin{tabular}{cccccccc} \hline
Model & \multicolumn{1}{l}{} & SVM \cite{cortes1995support} & TSVM \cite{4135685} & GBSVM \cite{xia2022gbsvm} & LSTSVM \cite{kumar2009least} & GBLSTSVM & LS-GBLSTSVM \\
 \multirow{2}{*}{Dataset}& \multicolumn{1}{c}{\multirow{2}{*}{Noise}} & ACC($\%$) & ACC($\%$) & ACC($\%$) & ACC($\%$) & ACC($\%$) & ACC($\%$) \\
 &  & ($c_1, \mu$) & ($c_1, c_2, \mu$) & ($c_1, \mu$) & ($c_1, c_2, \mu$) & ($c_1, c_2, \mu$) & ($c_1, c_2, \mu$) \\ \hline
\multirow{10}{*}{acute\_nephritis } & 0$\%$ & 41.67 & 90.00 & 65.58 & 41.66 & \textbf{100.00} & 94.44 \\
\multicolumn{1}{c}{\multirow{10}{*}{(120 $\times$ 6)}}
 & \multicolumn{1}{l}{} & (0.00001, 0.03125) & (0.00001, 0.00001, 0.03125) & (0.00001, 0.03125) & (0.00001, 0.00001, 0.03125) & (100000,  100000, 2) & (10, 1000, 2) \\
 & 5$\%$ & 41.67 & 90.00 & 65.59 & 41.67 & \textbf{100.00} & 83.33 \\
 & \multicolumn{1}{l}{} & (0.00001, 0.03125) & (0.001, 0.001, 0.03125) & (0.00001, 0.03125) & (0.00001, 0.00001, 0.03125) & (100000,  100000, 0.5) & (10, 1000, 2) \\
 & 10$\%$ & 41.67 & 90.00 & 84.44 & 41.67 & \textbf{100.00} & 91.67 \\
\hline
\end{tabular} }
\end{table*}

\begin{table*}[ht!]
 \caption*{\text{Table S.3:} (Continued)}
\label{tab:my-table70}
\centering
\resizebox{1.05\linewidth}{!}{
\begin{tabular}{cccccccc} \hline
Model & \multicolumn{1}{l}{} & SVM \cite{cortes1995support} & TSVM \cite{4135685} & GBSVM \cite{xia2022gbsvm} & LSTSVM \cite{kumar2009least} & GBLSTSVM & LS-GBLSTSVM \\
 \multirow{2}{*}{Dataset}& \multicolumn{1}{c}{\multirow{2}{*}{Noise}} & ACC($\%$) & ACC($\%$) & ACC($\%$) & ACC($\%$) & ACC($\%$) & ACC($\%$) \\
 &  & ($c_1, \mu$) & ($c_1, c_2, \mu$) & ($c_1, \mu$) & ($c_1, c_2, \mu$) & ($c_1, c_2, \mu$) & ($c_1, c_2, \mu$) \\ \hline

\multirow{10}{*}{ecoli-0-1-4-7\_vs\_2-3-5-6 } & 0$\%$ & 87.12 & \textbf{96.04} & 82.69 & 87.12 & 87.13 & 93.07 \\
 \multicolumn{1}{c}{\multirow{10}{*}{(336 $\times$7)}}
 & \multicolumn{1}{l}{} & (0.00001,  0.03125) & (0.1, 1, 0.5) & (100000,  8) & (0.00001, 0.00001, 0.03125) & (100000,  100000, 8) & (10, 1, 32) \\
 & 5$\%$ & 87.13 & 90.04 & 80.54 & 87.13 & \textbf{93.07} & 72.28 \\
 & \multicolumn{1}{l}{} & (0.00001,  0.03125) & (0.1, 0.1, 0.25) & (100000,  2) & (0.00001, 0.00001, 0.03125) & (100000,  100000, 32) & (0.00001, 0.01, 32) \\
 & 10$\%$ & 87.13 & 83.07 & 85.69 & 87.13 & \textbf{92.08} & 84.16 \\
\hline
\end{tabular} } 
\end{table*}

\begin{table*}[ht!]
 \caption*{\text{Table S.3:} (Continued)}
\label{tab:my-table80}
\centering
\resizebox{1.05\linewidth}{!}{
\begin{tabular}{cccccccc} \hline
Model & \multicolumn{1}{l}{} & SVM \cite{cortes1995support} & TSVM \cite{4135685} & GBSVM \cite{xia2022gbsvm} & LSTSVM \cite{kumar2009least} & GBLSTSVM & LS-GBLSTSVM \\
 \multirow{2}{*}{Dataset}& \multicolumn{1}{c}{\multirow{2}{*}{Noise}} & ACC($\%$) & ACC($\%$) & ACC($\%$) & ACC($\%$) & ACC($\%$) & ACC($\%$) \\
 &  & ($c_1, \mu$) & ($c_1, c_2, \mu$) & ($c_1, \mu$) & ($c_1, c_2, \mu$) & ($c_1, c_2, \mu$) & ($c_1, c_2, \mu$) \\ \hline

\multirow{10}{*}{haberman} & 0$\%$ & 81.52 & 75.35 & 57.61 & 80.43 & \textbf{82.61} & \textbf{82.61} \\
\multicolumn{1}{c}{\multirow{10}{*}{ (306 $\times$3)}}
 & \multicolumn{1}{l}{} & (0.1, 1) & (0.1, 1, 0.25) & (100000,  0.25) & (0.0001, 10, 0.125) & (100000,  100000, 32) & (100000,  100000, 32) \\
 & 5$\%$ & 81.52 & 70.26 & 72.10 & 80.43 & 80.43 & \textbf{82.61} \\
 & \multicolumn{1}{l}{} & (0.1, 1) & (0.1, 1, 0.125) & (100000,  0.25) & (0.0001, 10, 0.125) & (100000,  100000, 32) & (100000,  100000, 32) \\
 & 10$\%$ & \textbf{82.61} & 75.35 & 79.00 & 81.52 & 80.43 & 70.65 \\
\hline
\end{tabular} } 
\end{table*}

\begin{table*}[ht!]
 \caption*{\text{Table S.3:} (Continued)}
\label{tab:my-table90}
\centering
\resizebox{1.05\linewidth}{!}{
\begin{tabular}{cccccccc} \hline
Model & \multicolumn{1}{l}{} & SVM \cite{cortes1995support} & TSVM \cite{4135685} & GBSVM \cite{xia2022gbsvm} & LSTSVM \cite{kumar2009least} & GBLSTSVM & LS-GBLSTSVM \\
 \multirow{2}{*}{Dataset}& \multicolumn{1}{c}{\multirow{2}{*}{Noise}} & ACC($\%$) & ACC($\%$) & ACC($\%$) & ACC($\%$) & ACC($\%$) & ACC($\%$) \\
 &  & ($c_1, \mu$) & ($c_1, c_2, \mu$) & ($c_1, \mu$) & ($c_1, c_2, \mu$) & ($c_1, c_2, \mu$) & ($c_1, c_2, \mu$) \\ \hline
\multirow{10}{*}{ozone } & 0$\%$ & \textbf{96.58} & \textbf{96.58} & 80.00 & \textbf{96.58} & \textbf{96.58} & \textbf{96.58} \\
\multicolumn{1}{c}{\multirow{10}{*}{(2536 $\times$72)}}
 & \multicolumn{1}{l}{} & (0.00001,  0.03125) & (0.00001, 0.00001, 0.03125) & (100000,  32) & (0.00001, 0.00001, 0.03125) & (100000, 100000, 32) & (1, 0.1, 32) \\
 & 5$\%$ & \textbf{96.58} & 94.58 & 80.67 & \textbf{96.58} & \textbf{96.58} & \textbf{96.58} \\
 & \multicolumn{1}{l}{} & (0.00001,  0.03125) & (0.00001, 0.00001, 0.03125) & (100,  4) & (0.00001, 0.00001, 0.03125) & (100000, 100000, 32) & (1, 0.1, 32) \\
 & 10$\%$ & \textbf{96.58} & \textbf{96.58} & \textbf{96.58} & \textbf{96.58} & \textbf{96.58} & 90.54 \\
\hline
\end{tabular} }  
\end{table*}
\begin{table*}[ht!]
 \caption*{\text{Table S.3:} (Continued)}
\label{tab:my-table100}
\centering
\resizebox{1.05\linewidth}{!}{
\begin{tabular}{cccccccc} \hline
Model & \multicolumn{1}{l}{} & SVM \cite{cortes1995support} & TSVM \cite{4135685} & GBSVM \cite{xia2022gbsvm} & LSTSVM \cite{kumar2009least} & GBLSTSVM & LS-GBLSTSVM \\
 \multirow{2}{*}{Dataset}& \multicolumn{1}{c}{\multirow{2}{*}{Noise}} & ACC($\%$) & ACC($\%$) & ACC($\%$) & ACC($\%$) & ACC($\%$) & ACC($\%$) \\
 &  & ($c_1, \mu$) & ($c_1, c_2, \mu$) & ($c_1, \mu$) & ($c_1, c_2, \mu$) & ($c_1, c_2, \mu, pur.$) & ($c_1, c_2, \mu, pur.$) \\ \hline

\multirow{10}{*}{yeast-2\_vs\_4 } & 0$\%$ & 85.81 & 94.19 & \textbf{100.00} & 85.80 & 89.03 & 95.48 \\
\multicolumn{1}{c}{\multirow{10}{*}{ (514 $\times$8)}}
 & \multicolumn{1}{l}{} & (0.00001,  0.03125) & (0.1, 0.1, 0.03125) & (100000,  32) & (10, 0.00001, 0.03125) & (100000,  100000, 32) & (10000, 10, 32) \\
 & 5$\%$ & 85.81 & 94.19 & \textbf{95.48} & 85.81 & 85.81 & 85.81 \\
 & \multicolumn{1}{l}{} & (0.00001, 0.125) & (0.001, 0.01, 0.03125) & (100000, 32) & (10000, 0.00001, 0.03125) & (100000,  100000, 16) & (10, 1, 32) \\
 & 10$\%$ & 85.81 & 82.26 & 90.67 & 85.81 & \textbf{94.84} & 92.90 \\
\hline
\end{tabular}}  
\end{table*}

\renewcommand{\thetable}{S.4}
\begin{table}[ht!]
\centering
\caption{Classification accuracy of SLCE \cite{ghosh2024linear} across different noise levels on UCI and KEEL datasets with standard deviation.}
\label{tab:slce_noise_levels}
\resizebox{\textwidth}{!}{%
\begin{tabular}{|l|cc|cc|cc|cc|cc|}
\hline
 Noise & \multicolumn{2}{c|}{0\% } & \multicolumn{2}{c|}{5\% } & \multicolumn{2}{c|}{10\% } & \multicolumn{2}{c|}{15\% } & \multicolumn{2}{c|}{20\% } \\ \hline
                   Dataset   & Mean Acc& Std Dev& Mean Acc& Std Dev& Mean Acc & Std Dev & Mean Acc& Std Dev & Mean Acc & Std Dev \\ \hline
acute\_nephritis       & 100.00 & 0.00 & 96.67 & 3.51 & 91.33 & 5.17 & 87.89 & 5.02 & 64.33 & 9.83 \\ 
breast\_cancer         & 59.35 & 5.82 & 57.91 & 5.63 & 58.00 & 4.81 & 58.28 & 6.11 & 56.56 & 5.65 \\ 
breast\_cancer\_wisc\_prog & 67.53 & 5.27 & 64.47 & 5.61 & 60.87 & 6.98 & 61.93 & 6.03 & 58.13 & 6.47 \\ 
conn\_bench\_sonar\_mines\_rocks & 73.21 & 4.38 & 70.86 & 5.78 & 65.65 & 5.57 & 66.16 & 6.70 & 62.79 & 5.63 \\ 
crossplane130          & 99.49 & 1.03 & 96.10 & 3.33 & 90.56 & 4.68 & 85.23 & 11.36 & 58.67 & 12.19 \\ 
crossplane150          & 70.49 & 8.70 & 71.82 & 8.84 & 75.56 & 9.17 & 75.20 & 10.75 & 70.40 & 7.77 \\
\hline
\end{tabular}}
\end{table}

\end{document}